\documentclass[lettersize,journal]{IEEEtran}
\usepackage{amsmath,amsfonts}
\usepackage{array}
\usepackage[caption=false,font=normalsize,labelfont=sf,textfont=sf]{subfig}
\usepackage{textcomp}
\usepackage{stfloats}
\usepackage{url}
\usepackage{verbatim}
\usepackage{graphicx}
\usepackage[linesnumbered,ruled,vlined]{algorithm2e}
\usepackage{algcompatible}
\usepackage{makecell}
\usepackage[backend=bibtex]{biblatex} 
\usepackage{color}
\usepackage{ulem}
\usepackage{pifont}
\usepackage[export]{adjustbox}
\usepackage{xcolor}
\bibliography{mybibfile}

\begin{document}

\title{An efficient tangent based topologically distinctive path finding for grid maps}

\author{Zhuo Yao, Wei Wang$\ast$
\thanks{Corresponding by Wei Wang: wangweilab@buaa.edu.cn.
This work was supported by the National Key Research and Development Program of China under grant number 2020YFB1313600.}
}

\markboth{Journal of \LaTeX\ Class Files,~Vol.~14, No.~8, August~2021}%
{Shell \MakeLowercase{\textit{et al.}}: A Sample Article Using IEEEtran.cls for IEEE Journals}


\maketitle

\begin{abstract}
 

Conventional local planners frequently become trapped in a locally optimal trajectory, primarily due to their inability to traverse obstacles. Having a larger number of topologically distinctive paths increases the likelihood of finding the optimal trajectory. It is crucial to generate a substantial number of topologically distinctive paths in real-time. Accordingly, we propose an efficient path planning approach based on tangent graphs to yield multiple topologically distinctive paths. Diverging from existing algorithms, our method eliminates the necessity of distinguishing whether two paths belong to the same topology; instead, it generates multiple topologically distinctive paths based on the locally shortest property of tangents. Additionally, we introduce a priority constraint for the queue during graph search, thereby averting the exponential expansion of queue size. To illustrate the advantages of our method, we conducted a comparative analysis with various typical algorithms using a widely recognized public dataset\footnote{https://movingai.com/benchmarks/grids.html}. The results indicate that, on average, our method generates 320 topologically distinctive paths within a mere 100 milliseconds. This outcome underscores a significant enhancement in efficiency when compared to existing methods. To foster further research within the community, we have made the source code of our proposed algorithm publicly accessible\footnote{https://joeyao-bit.github.io/posts/2023/09/07/}. We anticipate that this framework will significantly contribute to the development of more efficient topologically distinctive path planning, along with related trajectory optimization and motion planning endeavors.

\end{abstract}


\section{Introduction}

As trajectory optimization and some motion planning algorithms take an initial path as input, and once a reference path is generated, it cannot be updated to belong to a different topology through gradient-based optimization. And the optimal trajectory under different constraints and requirements may belong to topologically distinctive paths, it is crucial to consider multiple topologically distinctive paths simultaneously.

Rosmann et al. \cite{rosmann2017integrated} proposed an integrated approach for efficient online trajectory planning of topologically distinctive mobile robot trajectories. Ranganeni et al. \cite{ranganeni2018effective} integrated user-defined homotopy constraints and footstep planning for humanoid robots, resulting in a significant speedup in planning, particularly in more complex scenarios. Palmieri et al. \cite{palmieri2015fast} introduced a Voronoi graph-based topology-distinctive path planning method to find multiple socially-aware paths, from which the robot can choose the best one based on a social cost function. Kim et al. \cite{kim2021topology} developed a perception-aware planner that selects the path with the highest perception quality from a set of multiple topologically distinctive paths for MAVs.

\begin{figure}[t] \scriptsize
\begin{minipage}{.48\linewidth}
  \centerline{\includegraphics[width=4.2cm, cframe=gray .2mm]{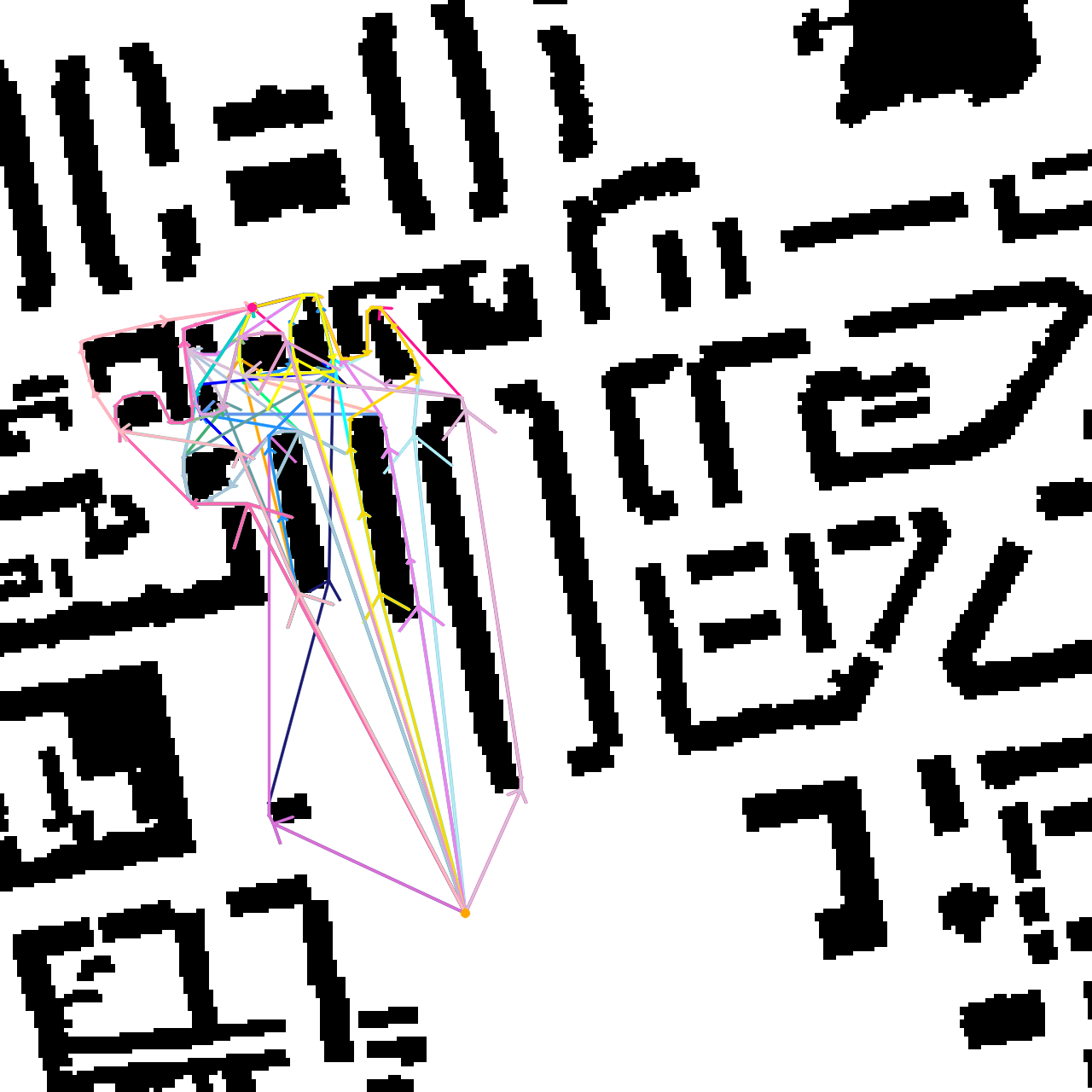}}
  \centerline{A: 50 paths, in 3.7ms}
\end{minipage}
\hfill
\begin{minipage}{.48\linewidth}
  \centerline{\includegraphics[width=4.2cm, cframe=gray 0.1mm]{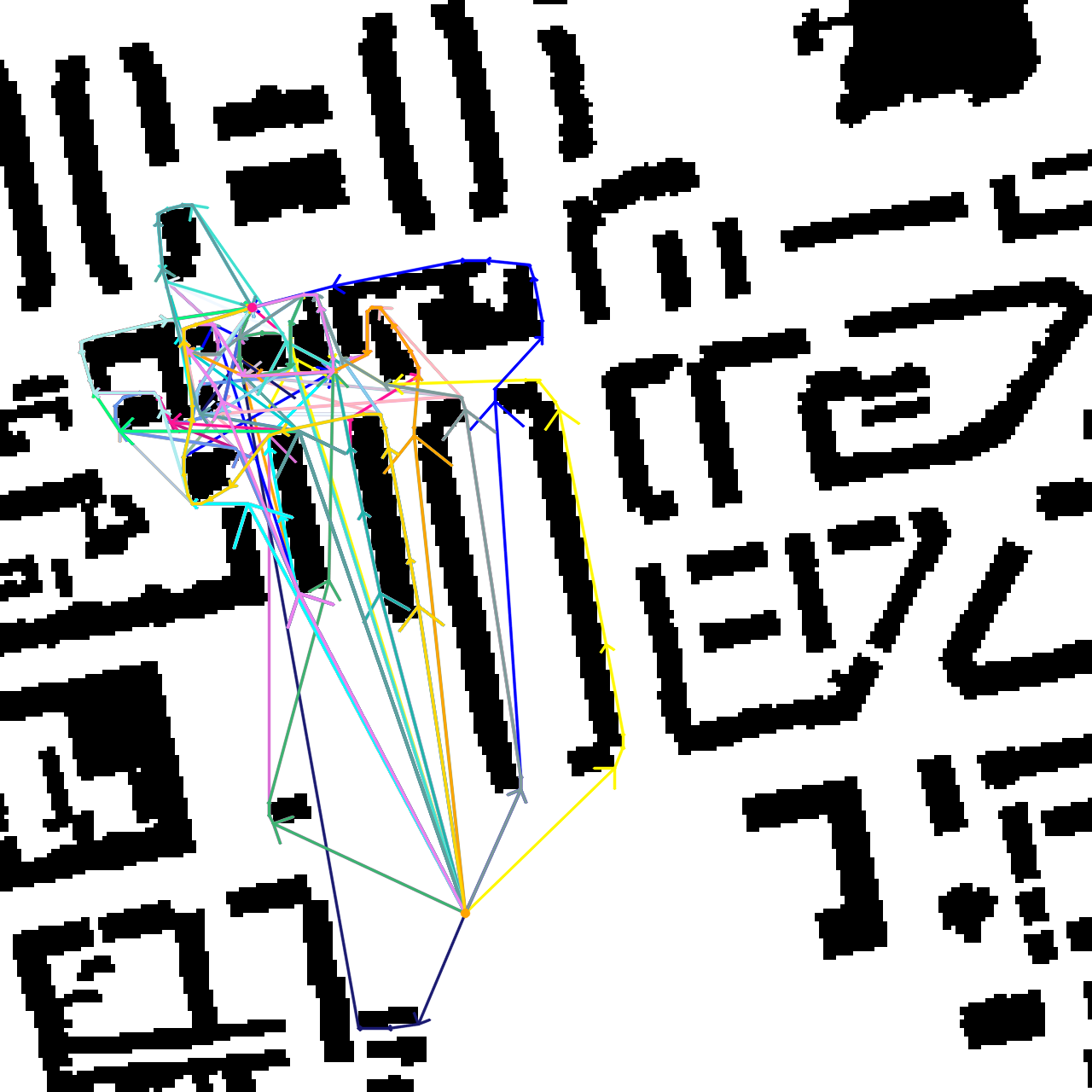}}
  \centerline{B: 100 paths, in 8.2ms}
\end{minipage}
\vfill
\begin{minipage}{.48\linewidth}
  \centerline{\includegraphics[width=4.2cm, cframe=gray 0.1mm]{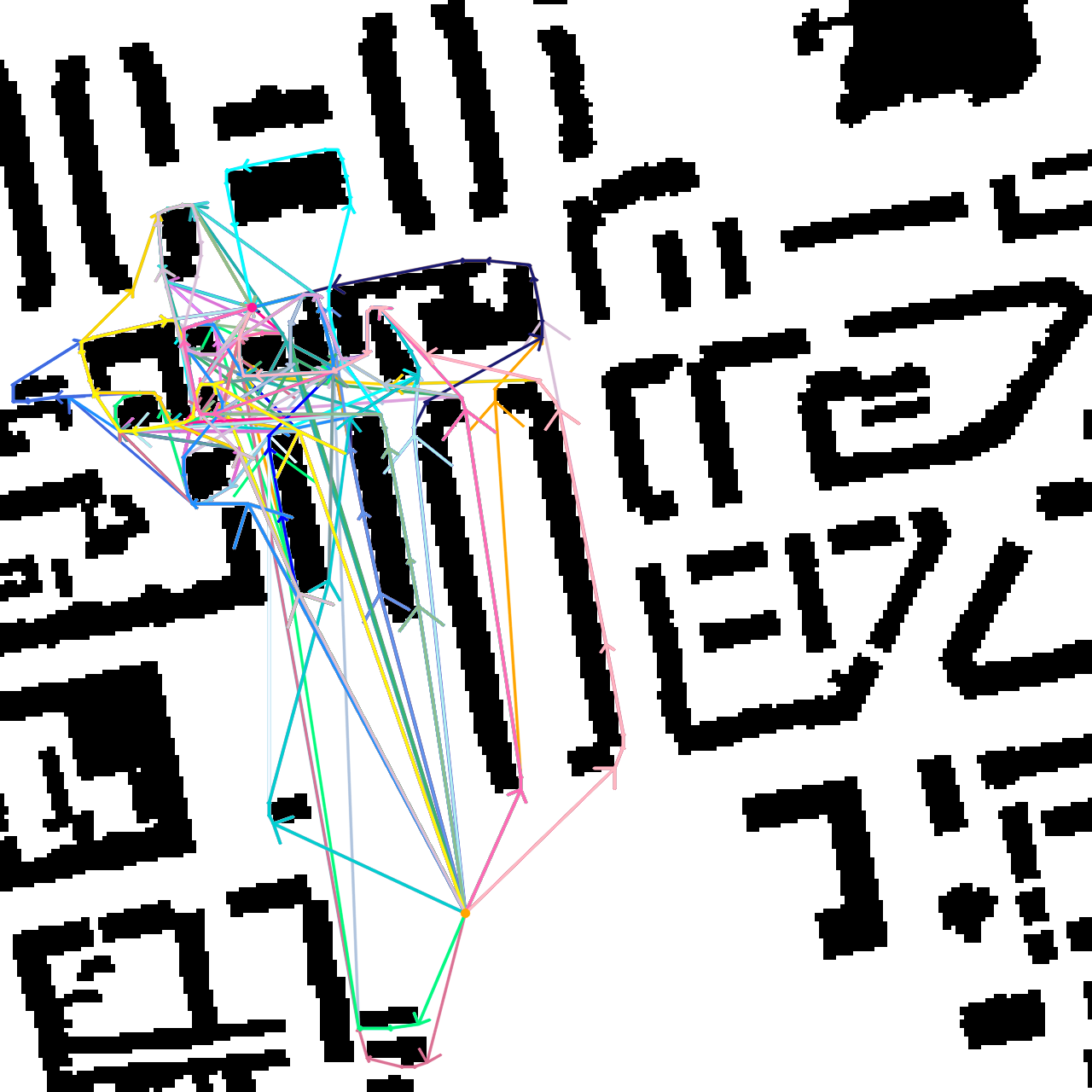}}
  \centerline{C: 200 paths, in 16.2ms}
\end{minipage}
\hfill
\begin{minipage}{.48\linewidth}
  \centerline{\includegraphics[width=4.2cm, cframe=gray 0.1mm]{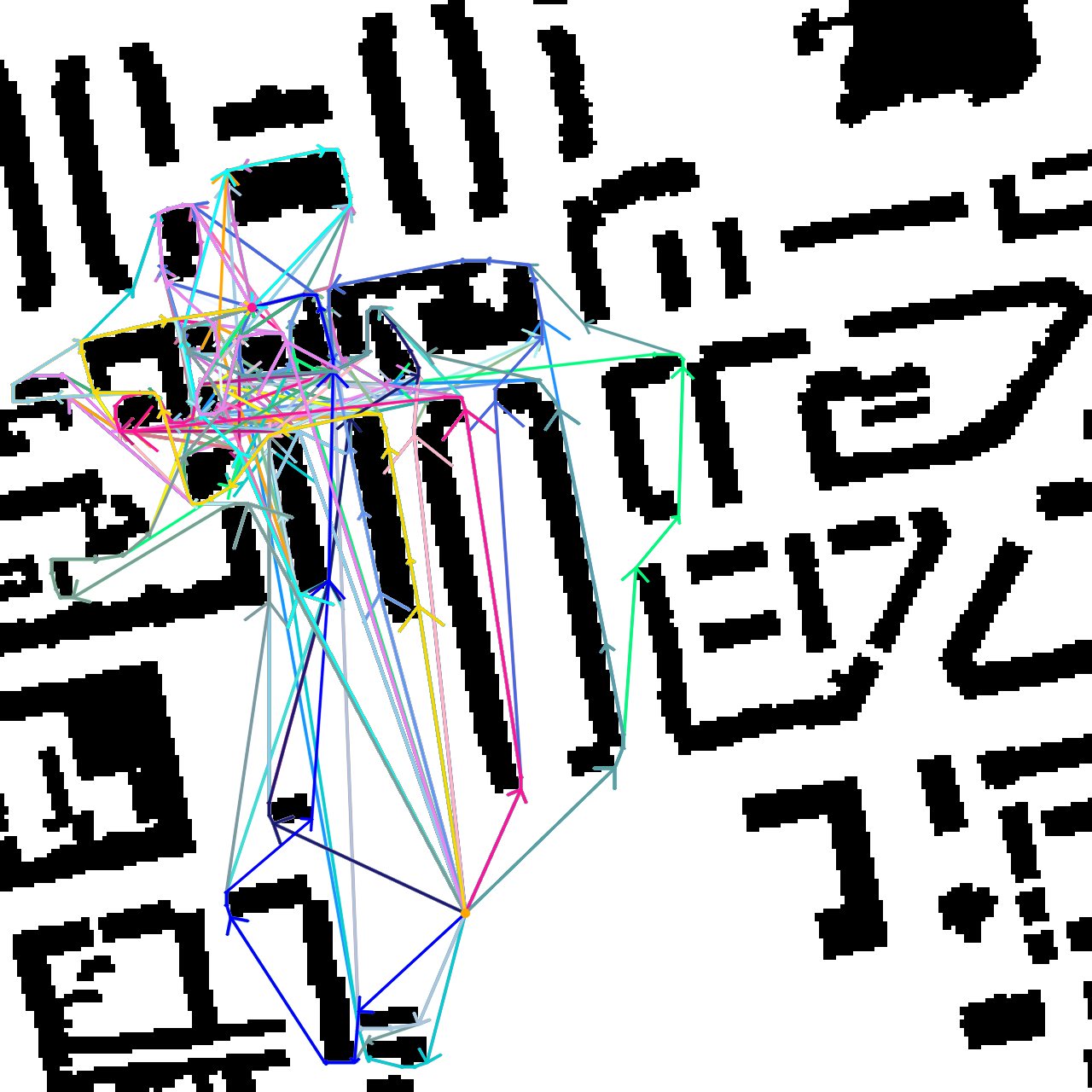}}
  \centerline{D: 400 paths, in 38.1ms}
\end{minipage}
\vfill
\caption{These figures display multiple paths, with some of them partially overlapping, between the same start and target points, all determined by our method. The map employed is a 256x256 grid map ("Berlin\_1\_256" from the mentioned public map dataset). The start and target points are highlighted in pink and yellow, respectively, with coordinates (59, 72) and (109, 214). These experiments were conducted on a standard laptop running the Ubuntu operating system, equipped with a 3.2GHz CPU and 16GB of memory. Further details can be found in the Results section. }
\label{examples}
\end{figure}


However, several difficulties are encountered by existing works. Firstly, some of them require the use of H-signature or similar indicators to determine whether two paths belong to the same topology. This results in an increase in time complexity as the number of paths to be considered grows, as it requires more calculations to determine if a new path and a previous path share the same topology. Secondly, for existing methods that rely on H-signature, the average time required to determine a path is directly related to the number of obstacles in the maps. Consequently, these methods become increasingly time-consuming when operating in environments with dense obstacles.

To address the limitations of H-signature and similar indicators, Voronoi-based approaches have been introduced. The distinctive property of Voronoi is that different sequences of Voronoi nodes inherently define topologically distinctive paths. Some of these methods have shown significant improvements in efficiency compared to algorithms that use H-signature. However, these methods still require a sequential search for topology-distinctive paths, one at a time.


To address the aforementioned issues, we propose a novel topologically distinctive path planning algorithm based on tangent graphs. This approach leverages the property that tangents form locally shortest paths, as discussed in previous works \cite{liu1991proposal, liu1992path, yao2019reinforcedrimjump}. Consequently, any two locally shortest paths inherently belong to different topologies. Our algorithm eliminates the need for indicators to determine path topology similarity and avoids the repetition of searches to obtain multiple paths. Instead, it retrieves all relevant paths in a single search, resulting in a significant improvement in efficiency when compared to existing methods. Additionally, we introduce a priority limitation mechanism to prevent the exponential growth of the queue size during graph search. The results demonstrate a remarkable enhancement in efficiency during the path planning process.A brief demonstration of our algorithm is presented in Fig. \ref{examples}.


The following sections of this article are organized as follows: section \ref{RelatedWork} offers an introduction to relevant studies on distinctive topology path planning; section \ref{Methodology} provides a detailed description of the key processes involved in our method; in section \ref{Results}, we delve into the details of the construction of the tangent graph on the specified maps. Additionally, we present a comprehensive comparison between our method and several typical methods in terms of time cost. This section also includes information about the mean time required to determine a path and how our method performs as the number of paths increases; finally, in Section \ref{Conclusion}, we discuss the results obtained and potential drawbacks of the proposed method.


This letter contributes the following:

\begin{enumerate}
    \item A topology-distinctive path planning method based on the locally shortest property of tangents.
    \item The introduction of a priority limitation mechanism to mitigate the exponential growth of queue size in path planning that utilizes breadth-first search.
\end{enumerate}

\section{Related works}
\label{RelatedWork}

Two trajectories $\tau_1$ and $\tau_2$, connecting the same start and end coordinates, are considered homotopic if one can be continuously deformed into the other without intersecting any obstacles. Otherwise, they are considered topologically distinctive. Based on whether they introduce an explicit indicator, existing topology-distinctive path planning methods can be categorized into two types.



The first type introduces an explicit indicator to determine whether two paths belong to the same topology, typically in relation to obstacles.

H-signature, proposed by Subhrajit Bhattacharya for 2-dimensional maps, is an indicator computed using the Cauchy integral theorem and the Residue theorem from complex analysis. It is defined as the integration of an ``obstacle marker function" along a path, where the obstacle marker function involves the representative point of each obstacle. Bhattacharya's work, as seen in references \cite{bhattacharya2010search, kim2012optimal, bhattacharya2012topological}, extends H-signature to 2D maps and later to 3D maps \cite{bhattacharya2012search}. Two paths share the same H-signature if and only if they belong to the same topology; otherwise, they do not. Combining the H-signature constraint with standard graph search algorithms like A*, these methods find multiple topology-distinctive paths by repeating graph searches multiple times. However, H-signature's efficiency is affected by several factors: 1, the more obstacles there are, the more calculations are needed to compute the H-signature of a path; 2, as the number of paths increases, the average time required to compute a single path increases linearly, as each path needs to be compared with the H-signature of existing paths, and the number of existing paths grows.

Markus Kuderer \cite{kuderer2014online} introduced the concept of the winding angle of a trajectory, which is defined as the sum of infinitesimal angle differences to the representative points of obstacles along the trajectory. Similar to H-signature, paths belonging to the same topology have the same winding angle, while those with distinctive topologies do not. To address the issue of multiple paths to the goal within the same homotopy class when using H-signature with A*, Kuderer and colleagues incorporated Voronoi graphs to ensure that each path corresponds to a unique homotopy class. In their results, as presented in \cite{kuderer2014online}, their approach showed a significant improvement in efficiency compared to using H-signature combined with A*.


TARRT* (Topology-Aware RRT*), as proposed by Yi et al. in \cite{yi2017topology}, introduces a method for determining the homotopic equivalence of two arbitrary paths based on properties of strings recognized by a Deterministic Finite Automaton (DFA). These strings also involve the representative point of each obstacle, with the condition that no point is allowed to lie on any line connecting any two other representative points. Specifically, non-obstacle regions are divided into a set of subregions by lines that intersect representative points, and the connections between these subregions are referred to as reference frames. TARRT* assigns a unique label to each reference frame and appends the label of the reference frames to the string whenever the path crosses them. This string-based approach appears to be more efficient than H-signature because its calculation doesn't involve all obstacle representative points. However, TARRT* shares a limitation with normal RRT (Rapidly Exploring Random Trees) in that they are both probabilistically complete methods for finding existing solutions. For instance, they may fail when the path must pass through a narrow, long opening in an obstacle.


The second type of methods does not rely on an explicit topology indicator.



As previously noted, path segments in the Voronoi graph are inherently unique between two obstacles, ensuring that different paths found in the Voronoi graph guarantee distinctive topology. Consequently, the Voronoi graph is widely employed in distinctive topological path planning.


Luigi Palmieri \cite{palmieri2015fast} introduced the Randomized Homotopy Classes Finder (RHCF), a fast randomized method that identifies a set of K paths lying in distinct homotopy classes using the Voronoi graph. The approach involves repeatedly searching for random paths in the Voronoi graph and saving them if they differ from all previous paths. The process stops when K different paths have been found. In their results, RHCF outperformed Yen's K shortest path finding method \cite{yen1971finding} in terms of speed.

In contrast to mobile robots, which require determining the optimal joint path in the task space, J.J. Rice \cite{rice2020multi} proposed a bifurcation branch algorithm to characterize the solution space with the bifurcation branch roadmap. This approach generates initial joint paths that identify all homotopy classes and locally optimal paths in all relevant homotopy classes with the lowest cost, which is very likely the globally optimal path. While this method is more capable than traditional approaches in finding the global optimal path, it faces challenges as the number of homotopy classes grows exponentially with the complexity of the solution space. This presents significant difficulties when extending it to topology-distinctive path planning for mobile robots.

\section{Methodology}
\label{Methodology}

In this section, we introduce the fundamental concepts utilized in our algorithm, which encompass the construction of the tangent graph and the utilization of graph search to identify multiple topology-distinctive paths. Additionally, we present a technique known as ``priority limitation," which is designed to prevent the exponential growth of the queue size during graph search.

\subsection{Definitions}

This section outlines the fundamental definitions utilized in our algorithms, covering key definitions related to tangents and constraints in path search.

\subsubsection{Grid space}

Let $\mathcal{C}_{\mathcal{N}}$ denote a finite $\mathcal{N}$-dimensional integer Euclidean space, where the size of the space is defined as $\mathcal{D}={d_{1}, d_{2},...,d_{i},...,d_\mathcal{N}}$ and $d_{i} \in \mathbb{N}$. The coordinate of an element $\textit{g}$ in this space is defined as a vector $(x_1,x_2,...,x_i,...,x_{\mathcal{N}})$, where $x_i \in ([0,d_i) \cap \mathbb{N})$. 

In the following, we manily focus on $\mathcal{C}_{2}$.

\subsubsection{Grid states}

There are only two possible states for a grid/element in $\mathcal{C}_{\mathcal{N}}$: passable or unpassable. The set of all passable grids in $\mathcal{C}_{\mathcal{N}}$ is denoted as $\mathcal{F} \rightarrow \mathcal{C}_{\mathcal{N}}$, while the set of all unpassable grids is denoted as $\mathcal{O} \rightarrow \mathcal{C}_{\mathcal{N}}$. Therefore, $(\mathcal{F} \rightarrow \mathcal{C}_{\mathcal{N}}) \cup (\mathcal{O} \rightarrow \mathcal{C}_{\mathcal{N}}) = \mathcal{C}_{\mathcal{N}}$.

For convenience, we denote all grids outside of $\mathcal{C}_{\mathcal{N}}$ as unpassable.

\subsubsection{Surface grid}

For a given grid $g \in \mathcal{C}_\mathcal{N}$, the on-obstacle-surface condition, also known as the surface grid, is defined as follows:

\begin{center}
$ \exists g' \in \delta_{f}(g), g' \in \mathcal{F} \rightarrow \mathcal{C}_{\mathcal{N}}$ $\textbf{and}$ $\exists g'' \in \delta_{f}(g), g'' \in \mathcal{O} \rightarrow \mathcal{C}_{\mathcal{N}} $ 
\end{center}

All surface grids of a grid space $\mathcal{C}_{\mathcal{N}}$ is denoted as $\mathcal{S} \rightarrow \mathcal{C}_{\mathcal{N}}$ . When build tangent graph, we take all surface grids as candidate of tangent nodes.

\subsubsection{Line-of-sight check}  

Denote the line connecting two grids $\textit{g}_{1}$ and $\textit{g}_{2}$ as $\textit{g}_{1} \rightarrow \textit{g}_{2}$. We define the set of all grids that $\textit{g}_{1} \rightarrow \textit{g}_{2}$ crosses as $\tau(\textit{g}_{1} \rightarrow \textit{g}_{2})$. $\textit{g}_{1} \rightarrow \textit{g}_{2}$ is said to be collided if $\exists g \in \tau(\textit{g}_{1} \rightarrow \textit{g}_{2}) , g \in \mathcal{O} \rightarrow \mathcal{C}_{\mathcal{N}}$. If ${\textit{g}}_{1} \rightarrow \textit{g}_{2}$ collides, it is denoted as $({\textit{g}}_{1} \rightarrow \textit{g}_{2}) \in \mathcal{O}$; otherwise, it is denoted as $({\textit{g}}_{1} \rightarrow \textit{g}_{2}) \in \mathcal{F}$.

\subsubsection{Distance metrics}

The distance between two grids ${\textit{g}}_{1}$ and $\textit{g}_{2}$ is denoted as $\Omega({\textit{g}}_{1} \rightarrow \textit{g}_{2})$, which is defined as the Euclidean distance between the two grids.

\subsubsection{Angle bewteen three grids}

Assuming there are three grids $g_1$, $g_2$, and $g_3$, the angle between $g_2 \rightarrow g_1$ and $g_2 \rightarrow g_3$ is denoted as $\theta(g_1, g_2, g_3)$.

\begin{center}
$\theta(g_1, g_2, g_3) = \arccos\left(\frac{(g_1 - g_2) \cdot (g_3 - g_2)}{\Vert g_1 - g_2 \Vert \cdot \Vert g_3 - g_2 \Vert}\right)$
\end{center}

\subsubsection{Neighborhood and frontier of grid}

\begin{figure}[h] \scriptsize 
\begin{minipage}{.23\linewidth} 
  \centerline{\includegraphics[width=1.8cm]{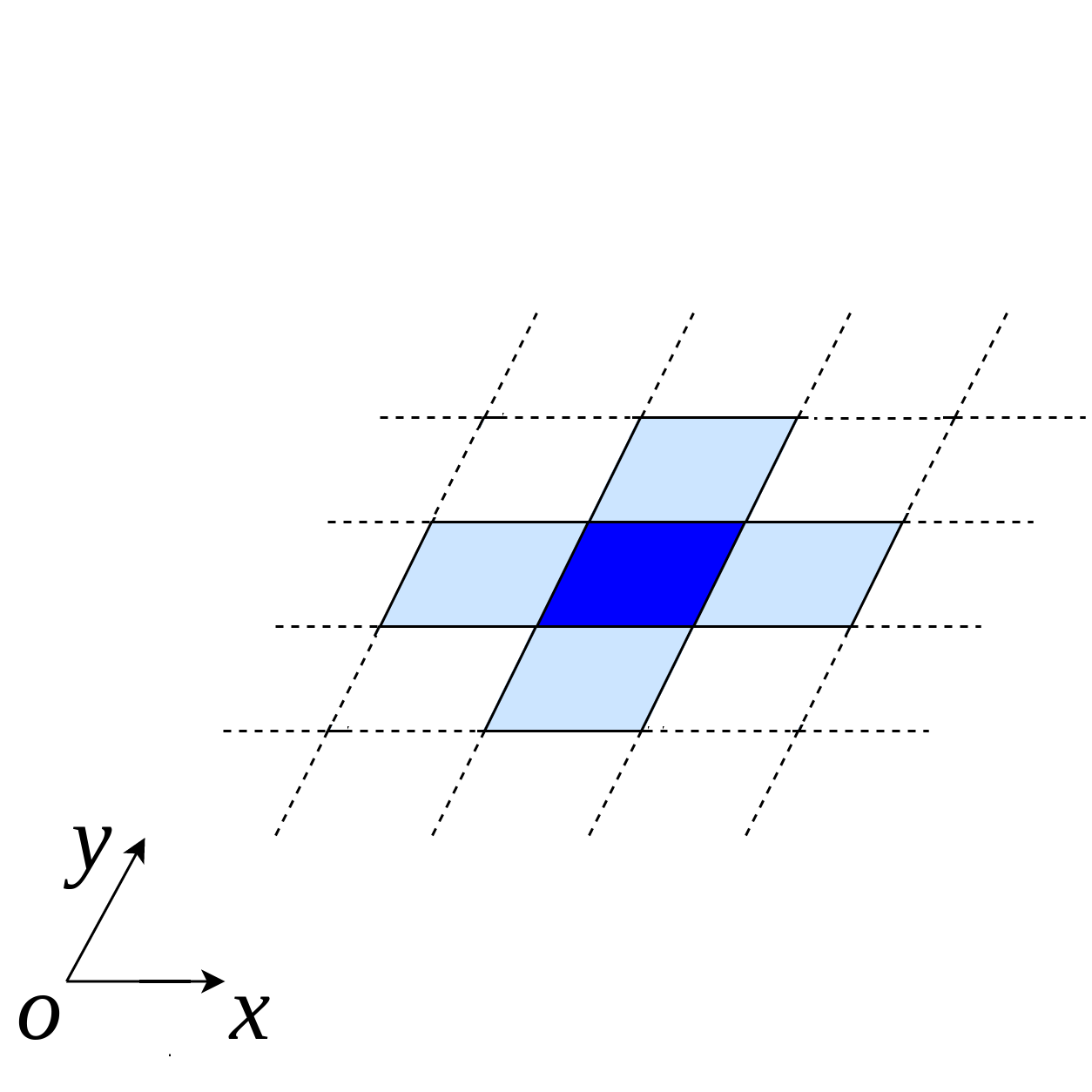}}
  \centerline{(A) $\delta(g)$ in $\mathcal{C}_2$ }

\end{minipage}
\hfill
\begin{minipage}{.23\linewidth}
  \centerline{\includegraphics[width=1.8cm]{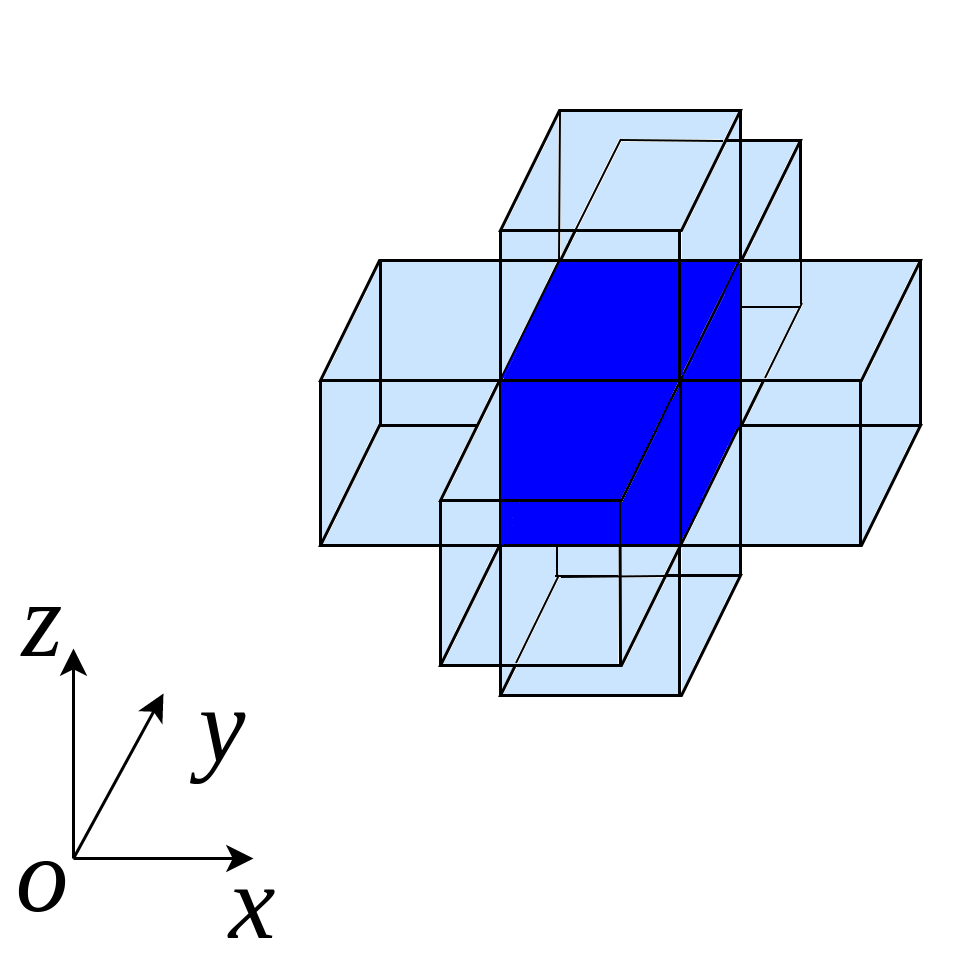}}
  \centerline{(B) $\delta(g)$ in $\mathcal{C}_3$ }
\end{minipage}
\hfill
\begin{minipage}{.23\linewidth}
  \centerline{\includegraphics[width=1.8cm]{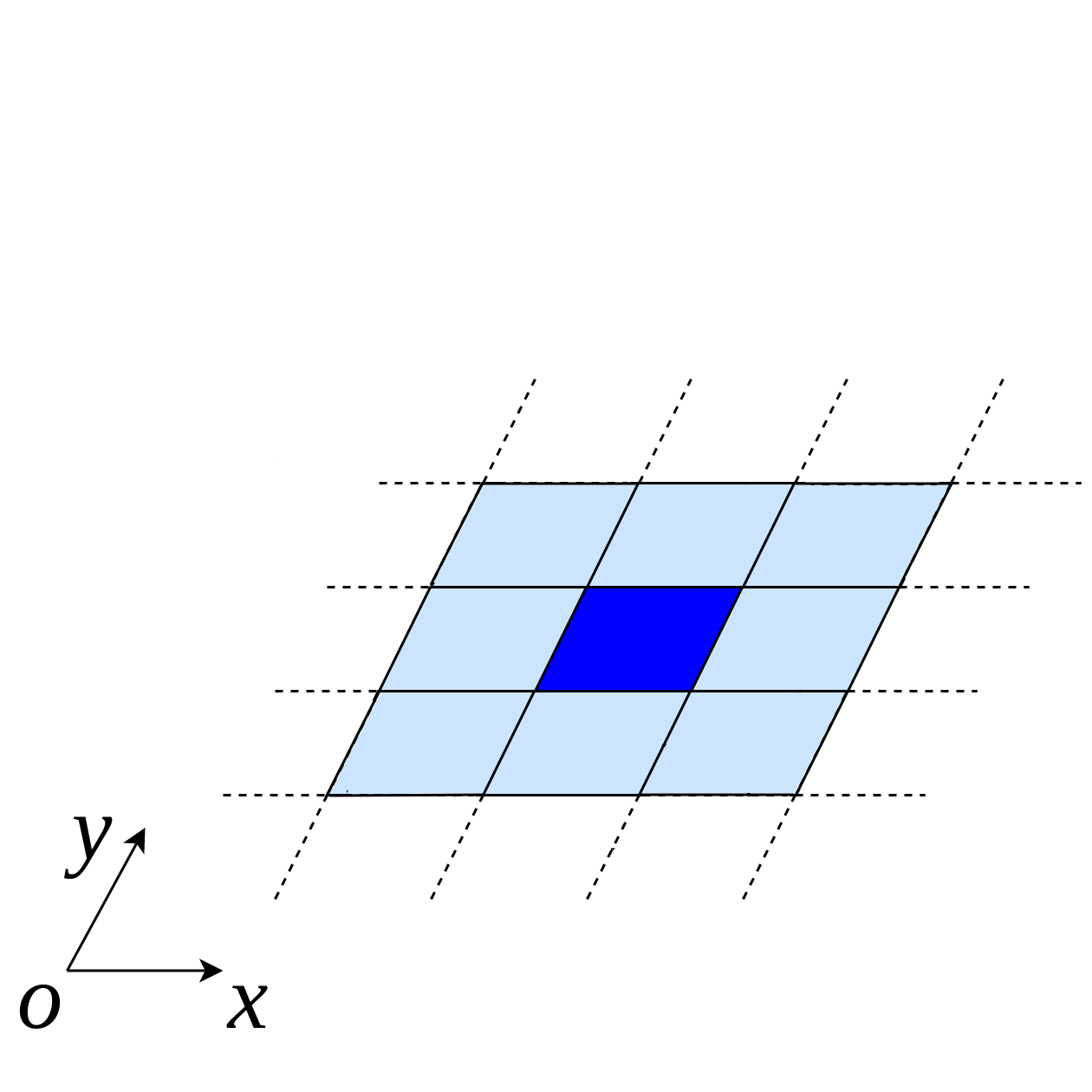}}
  \centerline{(C) ${\delta_f}(g)$ in $\mathcal{C}_2$ }
\end{minipage}
\hfill
\begin{minipage}{.23\linewidth}
  \centerline{\includegraphics[width=1.8cm]{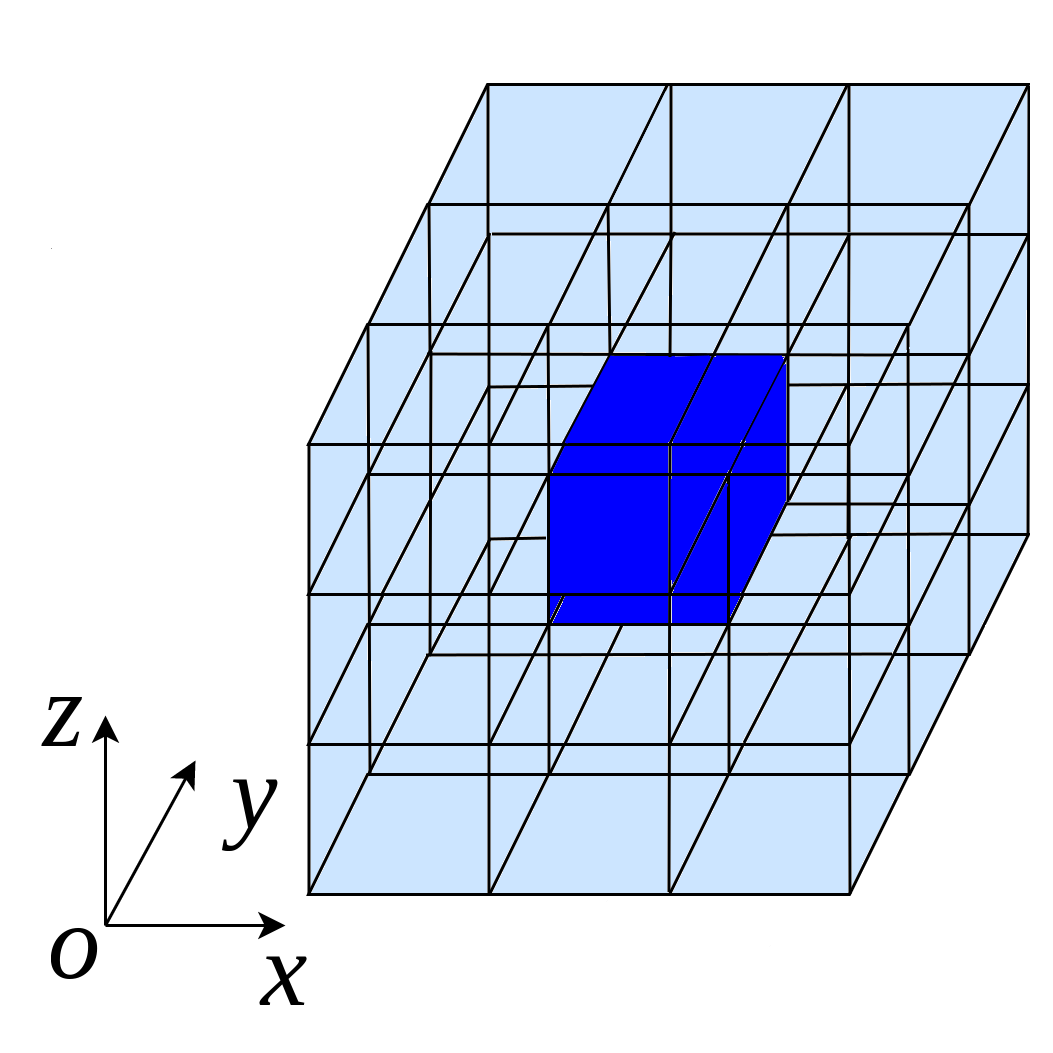}}
  \centerline{(D) ${\delta_f}(g)$ in $\mathcal{C}_3$ }
\end{minipage}
\vfill

\caption{Toy examples of neighborhood and frontier in $\mathcal{C}_{2}$ and $\mathcal{C}_{3}$ are shown in this figure, where $g$ is shown in deep blue, $\delta(g)$ and ${\delta_f}(g)$ are shown in light blue.}
\label{Neighborhood_and_frontier}
\end{figure}

Denote $\delta_f(g)$ as the frontier of grid $g$, which consists of the closest $3^\mathcal{N}$ grids near $g$. Toy examples of neighborhood and frontier in $\mathcal{C}_{2}$ and $\mathcal{C}_{3}$ are shown in Fig. \ref{Neighborhood_and_frontier}.

\subsubsection{Path} 

Path $p$ is defined as a sequence of waypoints. For a path, its first waypoint is the start, and if it is finished, its last waypoint is the target. All intermediate waypoints are nodes of the roadmap graph. A set of paths is defined as $P$. A finished path or a set of finished paths is defined as $p_f$ and $P_f$, respectively.

\subsubsection{Constraint of path search}

Path constraint pertains to the situation when a waypoint is added to an unfinished path, and determines what can be legally added to the path during path search. Path constraints can be categorized into three types, namely: point transfer constraints (PTC), edge transfer constraints (ETC), and iteration constraints (IC), as mentioned in Introduction.

Point transfer constraints (PTCs) dictate whether it is legal to establish a connection between two waypoints or from the start/target node to a waypoint. A commonly used example of a PTC is the line-of-sight check. Essentially, PTCs determine whether there exists an edge in the graph that connects two nodes. For convenience, we represent a single PTC as $\phi$ and a set of PTCs as $\Phi$.

Edge transfer constraints (ETCs) are used to determine the legality of forming a path through three waypoints. An example of an ETC is the taut constraint in ENL-SVG. For convenience, we denote a single ETC as $\psi$ and a set of ETCs as $\Psi$.

Iteration constraints (ICs) are constraints that determine the legality of more than three or all waypoints in a path. These constraints are only checked during the search process for forming a path in a roadmap graph. For example, the no-loop constraint is an IC used in distinctive topology path planning, while the global shortest constraint is used in shortest path planning. ICs are only determined during the path search process. For convenience, we denote a single IC as $\upsilon$ and a set of ICs as $\Upsilon$.

A potential question that may arise is why we note all constraints that involve more than three waypoints as ICs and do not include them in the precomputation stage. There are several reasons for this. Firstly, some constraints, such as the no loop constraint, can only be checked online during the path search process. Secondly, considering the time and storage cost, having more waypoints would require more time to check during the precomputation stage and would also require more storage space to save the result. Finally, we have not yet identified any crucial constraints that involve four or more waypoints and can be precomputed.

\subsection{Construct tangent graph} 

In this section, we present the methods for ensuring that the path is locally shortest and for constructing the tangent graph from surface grids in $\mathcal{C}_2$.

\subsubsection{Locally collide constraint}

\begin{figure}[h] \scriptsize
\begin{minipage}{.48\linewidth} 
  \centerline{\includegraphics[width=1.8cm]{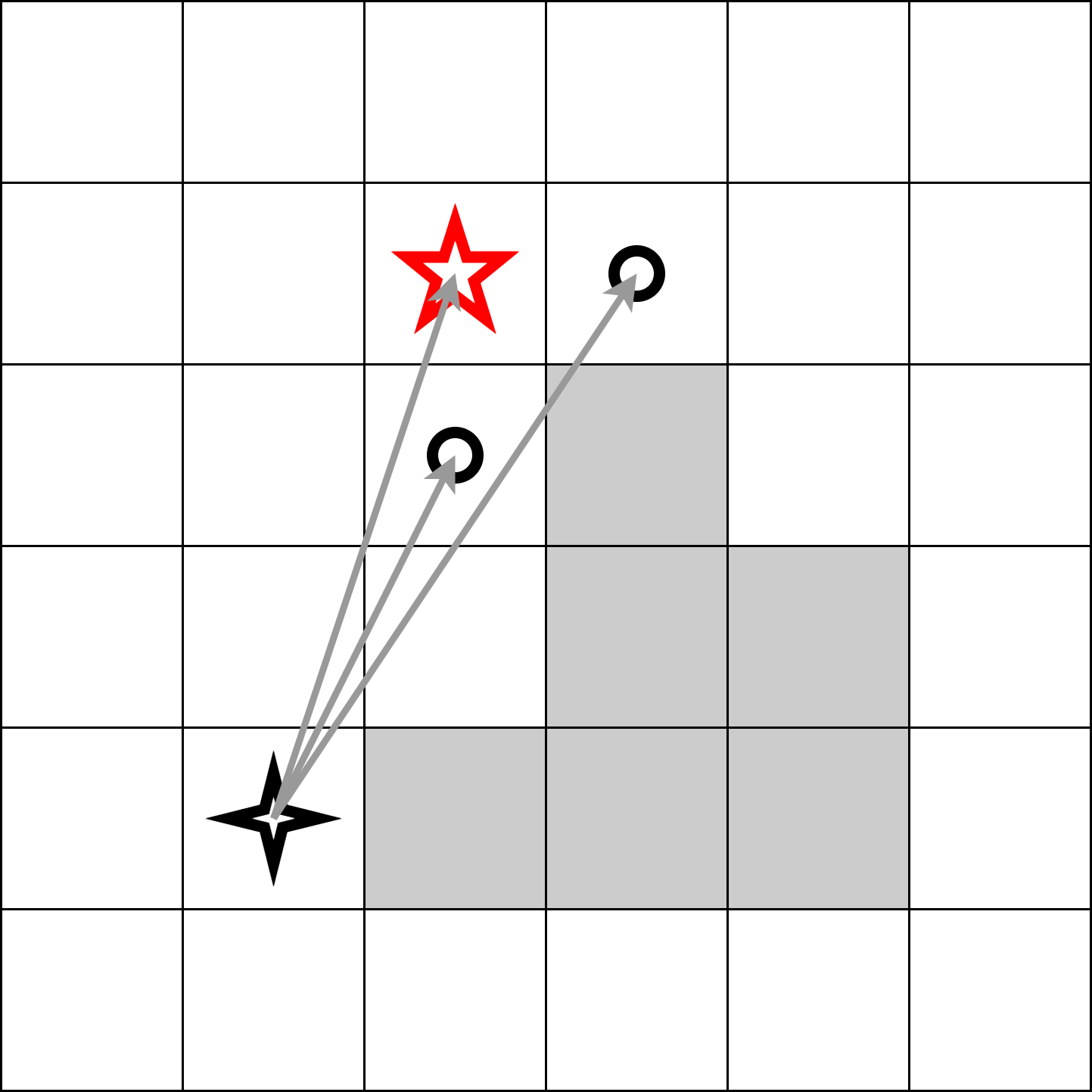}}
  \centerline{(A)}
\end{minipage}
\hfill
\begin{minipage}{.48\linewidth} 
  \centerline{\includegraphics[width=1.8cm]{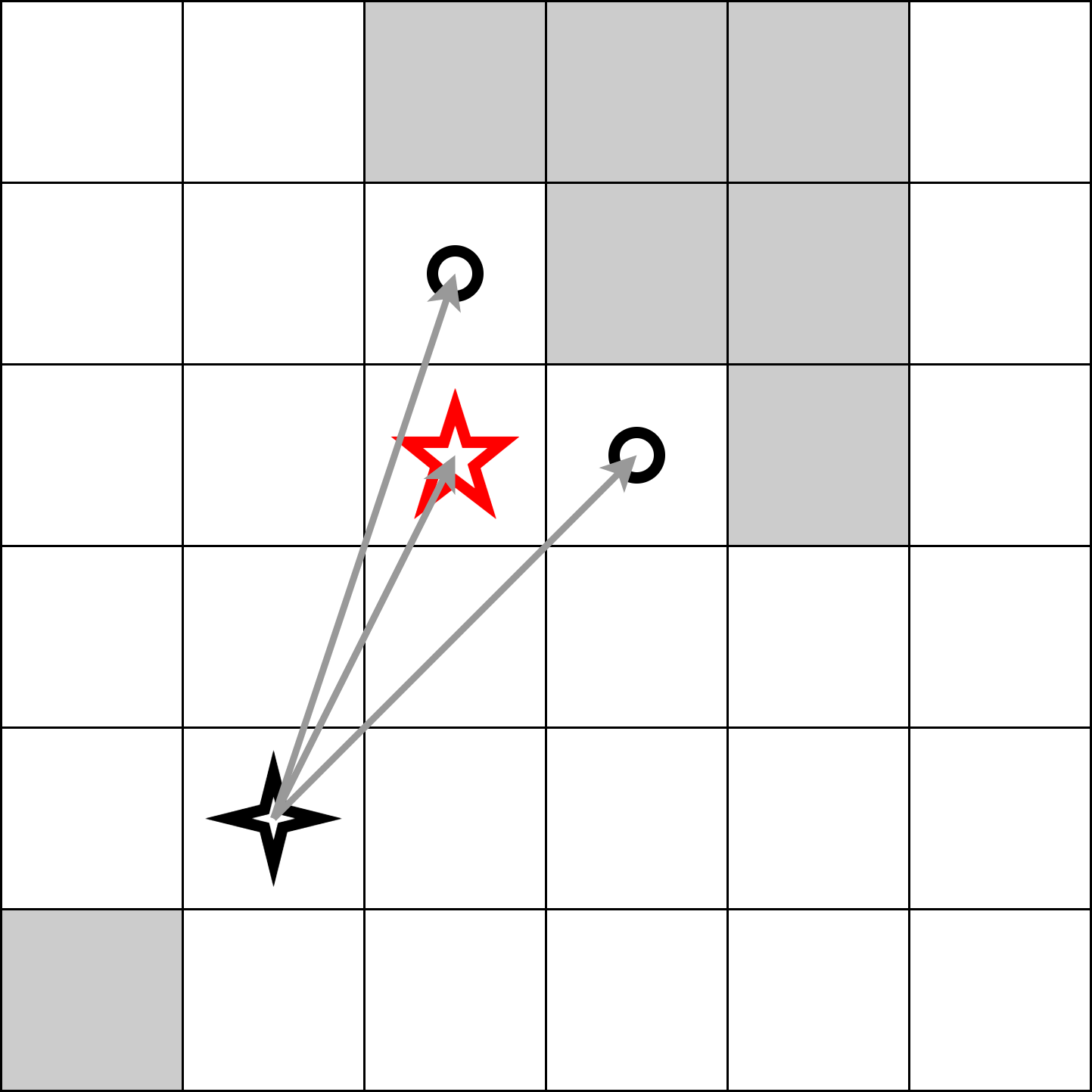}}
  \centerline{(B)}
\end{minipage}
\vfill
\caption{
The following figures depict two examples of node connections under $\mathcal{C}_{2}$, where (A) satisfies the local collide condition, while (B) does not. In these figures, four-pointed stars and pentagrams represent nodes, and the shadowed areas indicate obstacles.
}
\label{locally_collide}
\end{figure}

To ensure the shortest path, it is essential for each segment of the path to be locally shortest. For nodes located on the surface, it is required that there exist both a free connection and an occupied connection from their neighboring grids to another node. Furthermore, a collision connection should be longer than a free connection, as illustrated in Algorithm \ref{a3}. This condition is referred to as the ``local collide condition". Two nodes' connection is not locally shortest if they do not satisfy the ``local collide condition". Examples under $\mathcal{S}_2$ can be seen in Fig. \ref{locally_collide}. The locally collide constraint is a PTC that involves only two nodes. Therefore, it can be determined during the construction of the tangent graph, as described in Algorithm \ref{a4}.

\begin{algorithm}[h] 
 \normalem
\label{a3}
  \caption{Locally collide check}  
  \KwIn{$g_1, g_2,  \mathcal{C}_{\mathcal{N}}$}  
  \KwOut{True or False (whether $g_1 \rightarrow g_2$ is locally collide) }  
		\For{i = 1, 2} 
		{
		$\mathcal{D}_{\mathcal{O}}$ = 0; 
		$\mathcal{D}_{\mathcal{F}} = \infty;$
		$j = (i+1)\%2$; $//$ another grid  \\
		\If{$ \exists g' \in \delta_{f}(g_j),  g' \in \mathcal{O} \rightarrow  \mathcal{C}_{\mathcal{N}}$}
		{
		\For{$g' \in (\delta_{f}(g_j) \cap (\mathcal{S} \rightarrow \mathcal{C}_{\mathcal{N}})) $} 
		{
		\eIf{$(g_i \rightarrow g') \in \mathcal{O}$} 
		{\If{$ \mathcal{D}_{\mathcal{O}} < \Omega(g_i, g') $} {$\mathcal{D}_{\mathcal{O}} = \Omega(g_i, g')$;}}
		{\If{$ \mathcal{D}_{\mathcal{F}} > \Omega(g_i, g') $} {$\mathcal{D}_{\mathcal{F}} = \Omega(g_i, g')$;}}
		}
		\If{$\mathcal{D}_{\mathcal{O}} > {\mathcal{D}}_{\mathcal{F}}$} {return True;}
		}
		}	
		return False;	
\end{algorithm}

\subsubsection{Construct tangent graph}
When construct the tangent graph, we check whether every pairs of visible surface grids meet locally collide constraint, if meet, add the two grids as nodes of tangent graph and the connection as edge of tangent graph. The pseudocode for constructing the roadmap graph from the grid space is shown in Algorithm \ref{a4}. $V(g)$ means the grid that visible to grid $g$ and $\mathcal{G}(\mathcal{V}, \mathcal{E}) \rightarrow \mathcal{C}_{\mathcal{N}}$ means tangent graph of grid space $\mathcal{C}_{\mathcal{N}}$. For a map with sparse obstacles, we check the visibility of $g$ to all nodes in $\mathcal{V}$ to obtain $V(g)$. However, for maps with dense obstacles, we employ the Line-of-Sight Scan \cite{oh2017edge} to get $V(g)$, which is more efficient than performing vertex-to-vertex LOS checks.


Furthermore, to prevent the need for repeating these calculations every time the map is loaded, it is advisable to save the results to a file and load them during the framework loading process, thus saving time. To maximize the utility of memory space, we save the graph in binary file. 

\begin{algorithm}[h] 
 \normalem
\label{a4}
  \caption{Construct tangent graph}  
  \KwIn{$\mathcal{C}_{\mathcal{N}}$}
  \KwOut{
  $\mathcal{G}(\mathcal{V}, \mathcal{E}) \rightarrow \mathcal{C}_{\mathcal{N}}$
  }  
        $\mathcal{V} = \mathcal{S} \rightarrow \mathcal{C}_{\mathcal{N}}$; \\
		\For{$g \in \mathcal{V} $} 
		{
		\For{$g' \in V(g)$}
		{ 
		\If{$g \rightarrow g' $ meet locally collide constraint}
		{ 
		$\mathcal{V} \leftarrow g$; \\
		$\mathcal{E} \leftarrow (g \rightarrow g')$; 
		}
		}
		}		
		return $\mathcal{G}(\mathcal{V}, \mathcal{E})$;	
\end{algorithm}

\begin{figure*}[t]
\centering
\vspace*{8pt}
\includegraphics[width=12.5cm]{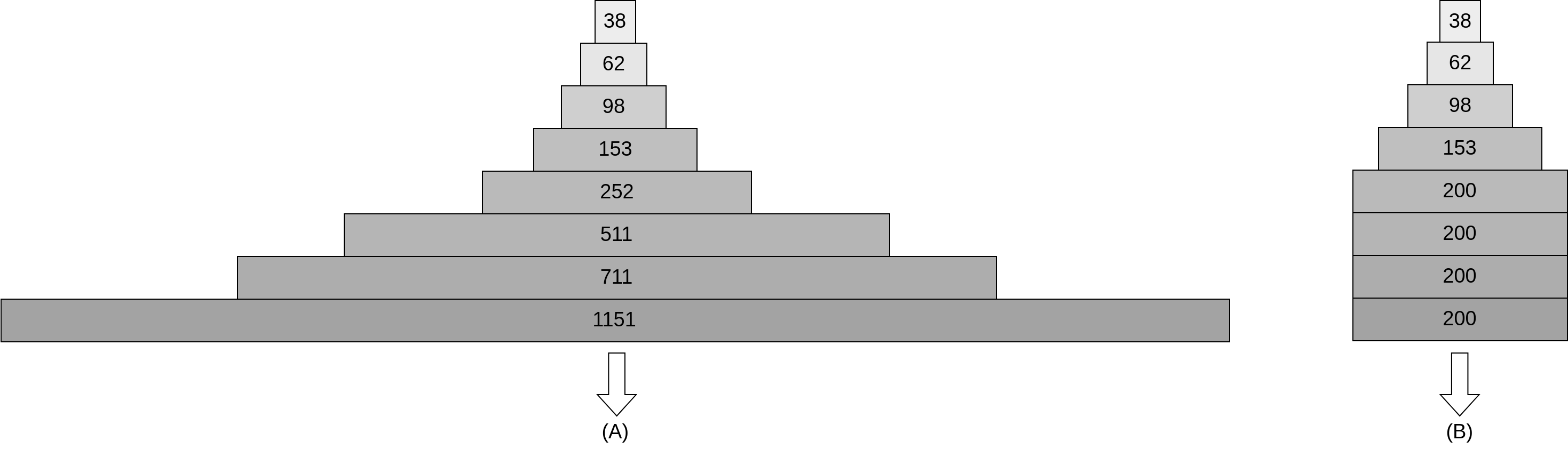}
\caption{
These figures illustrate how the queue size changes during BFS expansion before (Figure A) and after (Figure B) utilizing priority limitation in the same case (Fig. \ref{examples}) where 200 paths are being sought. As seen in Figures A and B, the queue size grows exponentially before the implementation of priority limitation but stops growing when it reaches 200 after the utilization of priority limitation. Consequently, before the introduction of priority limitation, it takes 48.5ms to obtain 200 paths, while it only takes 16.8ms after the implementation of priority limitation.
} 
\label{priority}
\end{figure*}

\subsection{Path search}

\subsubsection{No loop constraint}
As mentioned in the related literature, path planning with distinctive topology does not involve loops. In this context, loops can take two forms: either a waypoint is visited repeatedly within a single path, or the path intersects with itself. It is crucial to account for these scenarios when attempting to find multiple paths. It's worth noting that as the number of waypoints in a path increases, the time required for loop detection also increases. Experimental results indicate that the ratio of ``no loop constraint check" in the total time cost increases as the required number of paths increases.

\subsubsection{Geting closer to obstacle}
To ensure a locally shortest edge transfer, there exists a grid $g' \in ({\delta}_{f}(g_2) \cap \mathcal{O} \rightarrow \mathcal{S}_{\mathcal{N}})$ in the cone formed by $g_1 \rightarrow g_2 \rightarrow g_3$, including the obstacle grids near the frontier of $g$. Otherwise, it is feasible to construct a shorter locally path. The algorithm for this process is shown in Algorithm \ref{a6}. $g_c$ means a point located at the center of the cone formed by rotating $g_1 \rightarrow g_2 \rightarrow g_3$.

\begin{algorithm}[h] 
 \normalem
\label{a6}
  \caption{Geting closer to obstacle}  
  \KwIn{$g_1, g_2, g_3, \mathcal{C}_{\mathcal{N}}$}  
  \KwOut{True or False}
  		$\theta_{c} = \theta(g_1, g_2, g_3)/2;$ \\  
  		$ g_c = g_2 + \frac{g_1 - g_2}{||g_1 - g_2||} + \frac{g_3 - g_2}{||g_3 - g_2||}$; \\
		\For{ $g' \in ({\delta}_{f}(g_2) \cap \mathcal{O} \rightarrow \mathcal{S}_{\mathcal{N}} )$}
		{
		\If{$ \theta(g_1, g_c, g') < \theta_c $ } {
			return True;		
		}		
		}	
		return False;	
\end{algorithm}

\begin{figure}[h]
\centering
\vspace*{8pt}
\includegraphics[width=2.5cm]{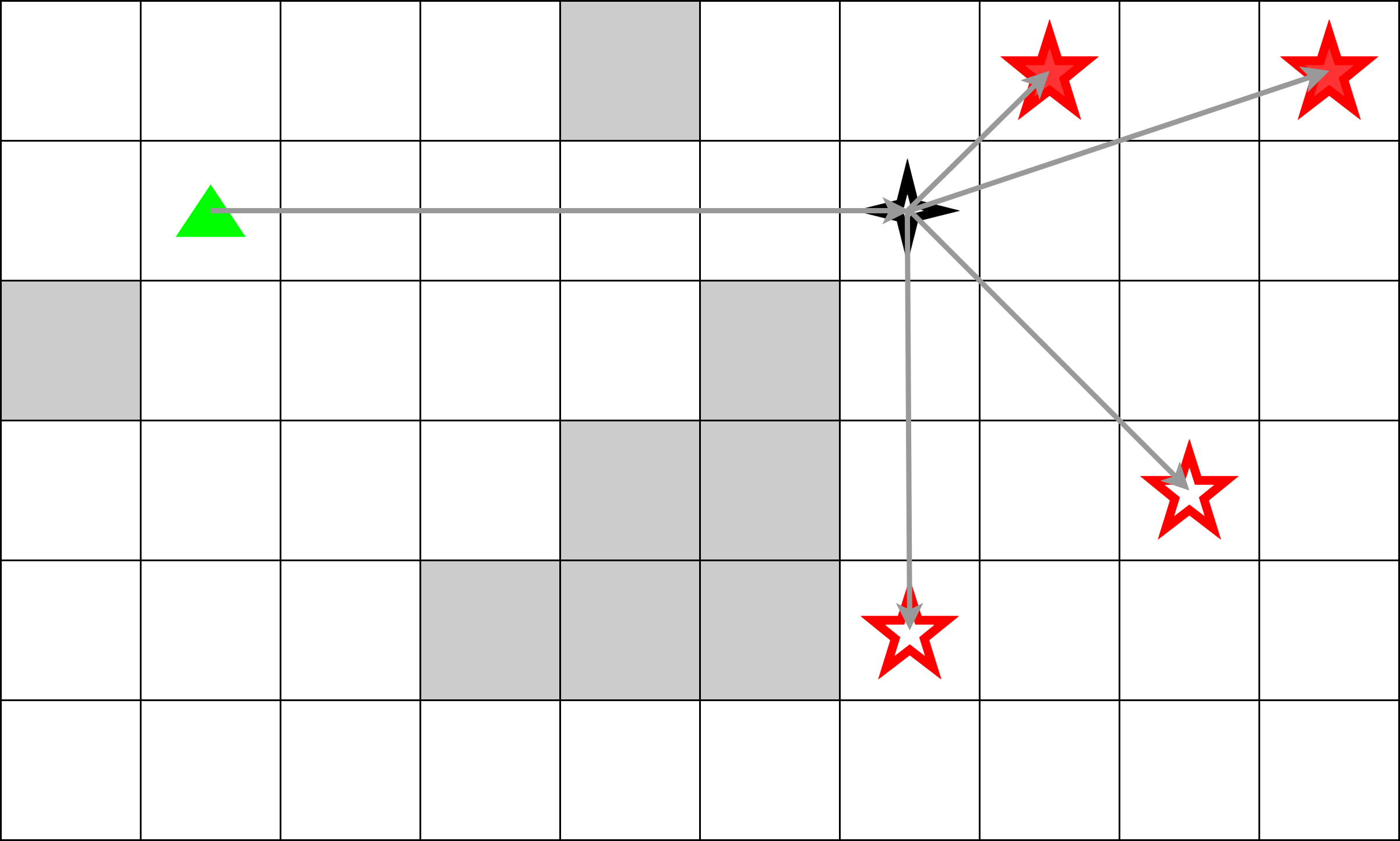}
\caption{
This figure displays two cases of edge transitions where the path approaches an obstacle (represented by the hollow pentagrams) and two cases where the edges do not approach obstacles (represented by the filled pentagrams). The initial edge $g_1 \rightarrow g_2$ is depicted as a triangle, and the endpoint $g_2$ is represented by a cross star. The four cases for the next edge $g_2 \rightarrow g_3$ are illustrated using pentagrams.
} 
\label{get_closer_obstacle}
\end{figure}

Examples of paths that get closer to an obstacle under constraint $\mathcal{C}_2$ are depicted in Fig. \ref{get_closer_obstacle}.

\subsubsection{Graph search}

\begin{algorithm}[t] 
 \normalem
\label{a8}
  \caption{Create initial paths}  
  \KwIn{$g_s, g_t, \Phi, \mathcal{C}_\mathcal{N}, \mathcal{G}(\mathcal{V},\mathcal{E})$ }  
  \KwOut{$E_i$ }
  $E_i = \emptyset;$ \\
  $\mathcal{V} \leftarrow g_s$;\\
  $\mathcal{V} \leftarrow g_t$;\\
  \For{$g' \in (V(g_s) \cap \mathcal{V})$} 
  {
  \If{$(g_s \rightarrow g')$ not meet all $\Phi$} {
  continue;
  }
  $\mathcal{E} \leftarrow (g_s \rightarrow g')$; \\  
  $E_i \leftarrow (g_s \rightarrow g_1)$;
  }
  \For{$g' \in (V(g_t) \cap \mathcal{V})$} 
  {
  \If{$(g' \rightarrow g_t)$ not meet all $\Phi$} {
  continue;
  }
  $\mathcal{E} \leftarrow (g' \rightarrow g_t)$; \\  
  }
  return $E_i$;
\end{algorithm}


Since the start and target points are typically not nodes of $\mathcal{G}$, it is necessary to establish connections from the start ($g_s$) and target point ($g_t$) to the tangent graph. This process consists of two sections: 1, establishing a connection between the start/target points and $\mathcal{V}$ of $\mathcal{G}(\mathcal{V}, \mathcal{E})$ to obtain initial edges $E_i$ that meet the requirements of the PTC (locally collide constraint); 2, adding the new edges that satisfy the edge transfer constraint ("Getting closer to obstacle") to $\mathcal{G}(\mathcal{V}, \mathcal{E})$; 3, creating initial paths that contain the first edge. This process results in the set of first edges for the incomplete paths, i.e., the initial paths.


The pseudo-code for creating initial paths is presented in Algorithm \ref{a8}. In the algorithm, $V(g_s) \cap \mathcal{V}$ and $V(g_t) \cap \mathcal{V}$ represent nodes in the tangent graph that are visible from the start and target points, respectively.  In lines 4 to 8, connections are established from the start to $\mathcal{G}(\mathcal{V}, \mathcal{E})$, while in lines 9 to 12, connections are established from the target to $\mathcal{G}(\mathcal{V}, \mathcal{E})$.



After establishing connections from the start/target to the tangent graph, we employ Breadth-First Search (BFS) to determine multiple distinctive topology paths during a single search. During the path search, we consider ETCs and ICs, while PTCs are not considered, as we have already addressed them during the creation of initial paths and the construction of the tangent graph.

As mentioned earlier, we incorporate the no loop constraint (IC) and the ``getting closer to obstacle" constraint (ETC) to obtain all locally shortest paths, which are also distinctive topology paths. Therefore, BFS is capable of finding multiple distinctive topology paths because each path found in the tangent graph is locally shortest and has distinctive topology when compared to others.

However, finding all distinctive topology paths can be time-consuming, especially in the case of large-scale maps. To address this, we have updated BFS to stop searching when the count of finished paths reaches a required number ($K$). Additionally, our BFS will exit when all paths reach the target, as the total number of distinctive topology paths may be less than the required value in some cases. An early version of finding distinctive topology paths via locally shortest paths was described in our PrePrint article \cite{yao2021tangent}.

\begin{algorithm}[t] 
 \normalem
\label{a11}
  \caption{Adapted breadth first search with priority}
  \KwIn{$g_t, E_i, \mathcal{G}(\mathcal{V}, \mathcal{E}), \Psi, \Upsilon, K$}  
  \KwOut{$P_f$}
  $P_u = \emptyset$; $//$ final paths \\ 
  \For{$e=(g_s \rightarrow g1) \in E_i$} {
  $p_u = \left\lbrace e \right\rbrace $; \\
  $P_u \leftarrow p$;
  }
  $P'_u = \emptyset $; \\
  $P_f = \emptyset $; \\
  \textbf{$P_{secondary} = \emptyset$;} \\
  \While{$P_u \neq \emptyset$ $\And$ not all (or have $K$) paths in $P_u$ reach target } {
  \For{$p_c \in P_u$ } {  
  $e_l$ = last edge of $p_c$; \\
  subsequent\_edges = edges of $\mathcal{E}$  that meet all $\Psi$ if it transfer to $e_l$; \\
  \For{e $\in$ subsequent edges} { 
  \If{$e_l \rightarrow e$ not meet all $\Psi$} {
    continue; \\
  }
  \If{$(p_c, e)$ not meet all $\Upsilon$} { continue; }
  \If{$e$ connect to $g_t$} {
  $P_f = p_u \leftarrow e$; \\
  }
  $p'_u = p_u \leftarrow e$; \\
  $P'_u \leftarrow p'_u$; \\
  }
  }
  \textbf{$P_u$ = pop top $K$ paths in $P'_u$;} \\
  \eIf{\textbf{$P_u$ have $K$ paths}} {
  \textbf{$P_{secondary} \leftarrow P'_u $}; \\  
  \textbf{sort $P_{secondary}$ in order of increasing heuristic value}; \\
  } {
  \textbf{$P_u \leftarrow$ pop paths in $P_{secondary}$ till $P_{secondary} = \emptyset$ or $P_u$ have $K$ paths;}  
  }
  }
  return $P_f$;
\end{algorithm}

\subsubsection{Priority limitation during graph search}

One drawback of the adapted BFS is the exponential growth in the size of unfinished paths ($P_u$ in Algorithm \ref{a11}) as the number of expansions increases. To mitigate this issue, we introduce a priority queue to limit the size of unfinished paths during each iteration to $K$. More specifically, we select the top $K$ paths with the minimal heuristic value (calculated in the same way as A*) as candidates for expansion. The remaining unfinished paths are cached in a secondary queue, sorted by their heuristic values.

\begin{figure}[t] \scriptsize
\begin{minipage}{.24\linewidth}
  \centerline{\includegraphics[width=2.1cm]{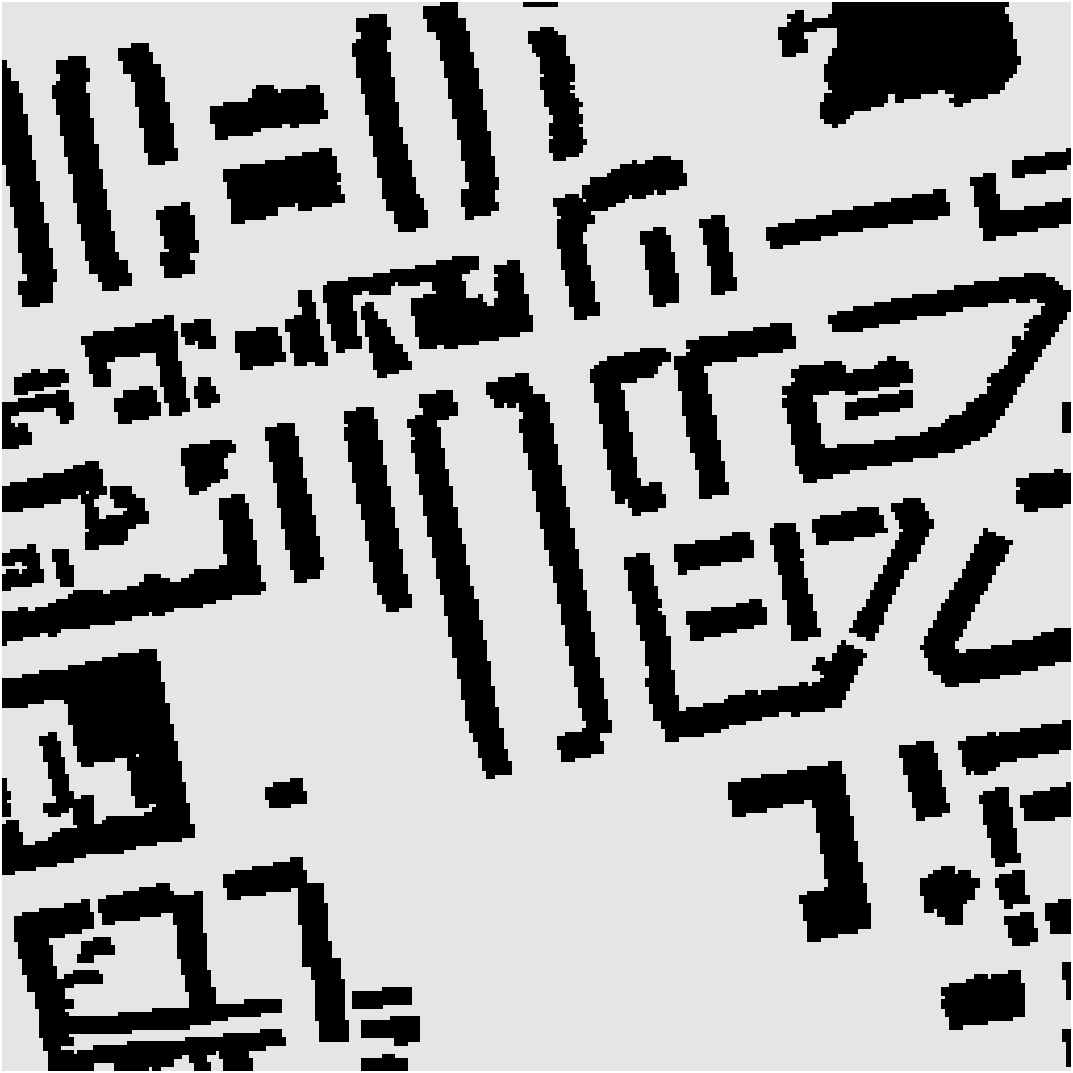}}
  \centerline{A: Berlin\_1\_256}
  \centerline{256*256}
\end{minipage}
\hfill
\begin{minipage}{.24\linewidth}
  \centerline{\includegraphics[width=2.1cm]{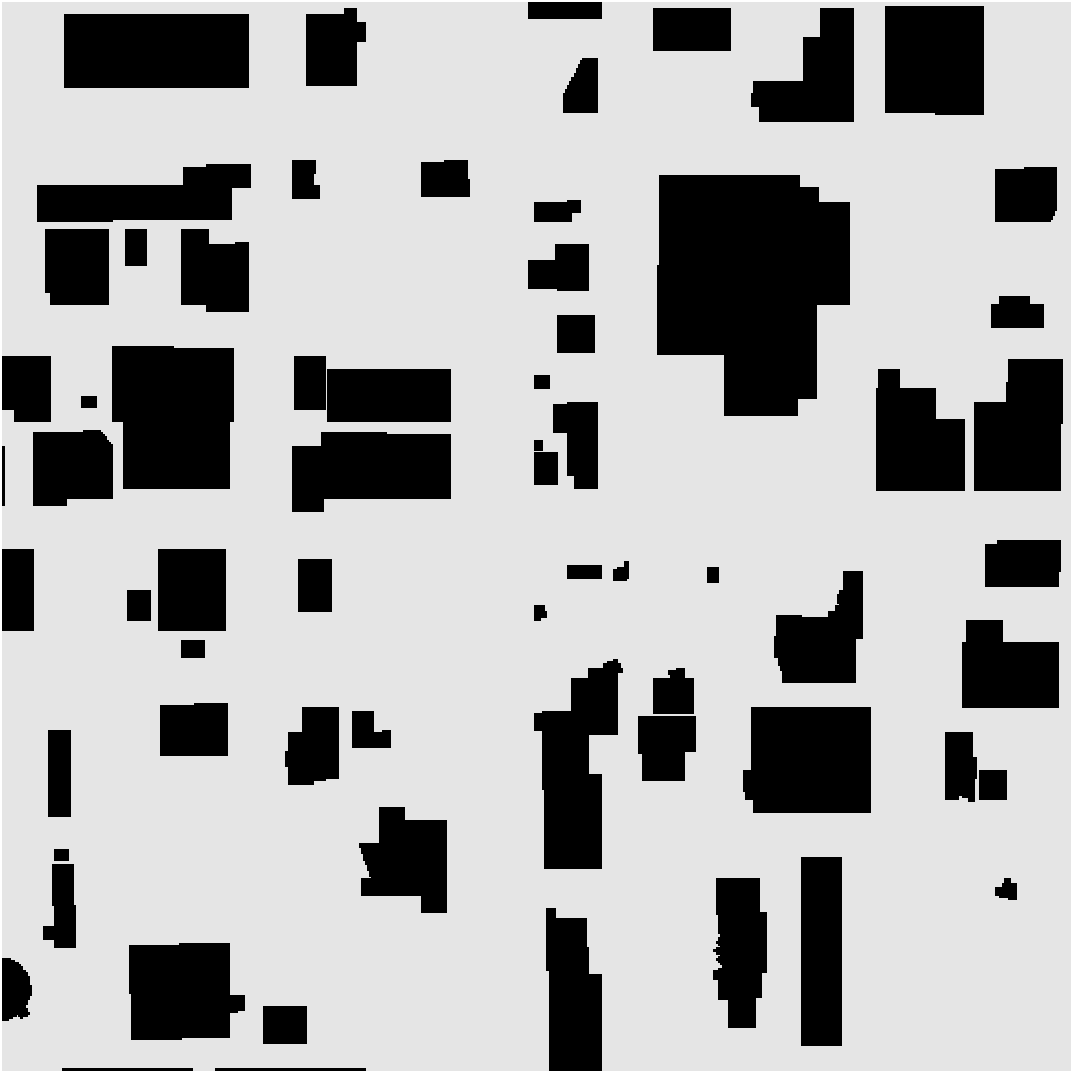}}
  \centerline{B: Denver\_2\_512}
  \centerline{512*512}
\end{minipage}
\hfill
\begin{minipage}{.24\linewidth}
  \centerline{\includegraphics[width=2.1cm]{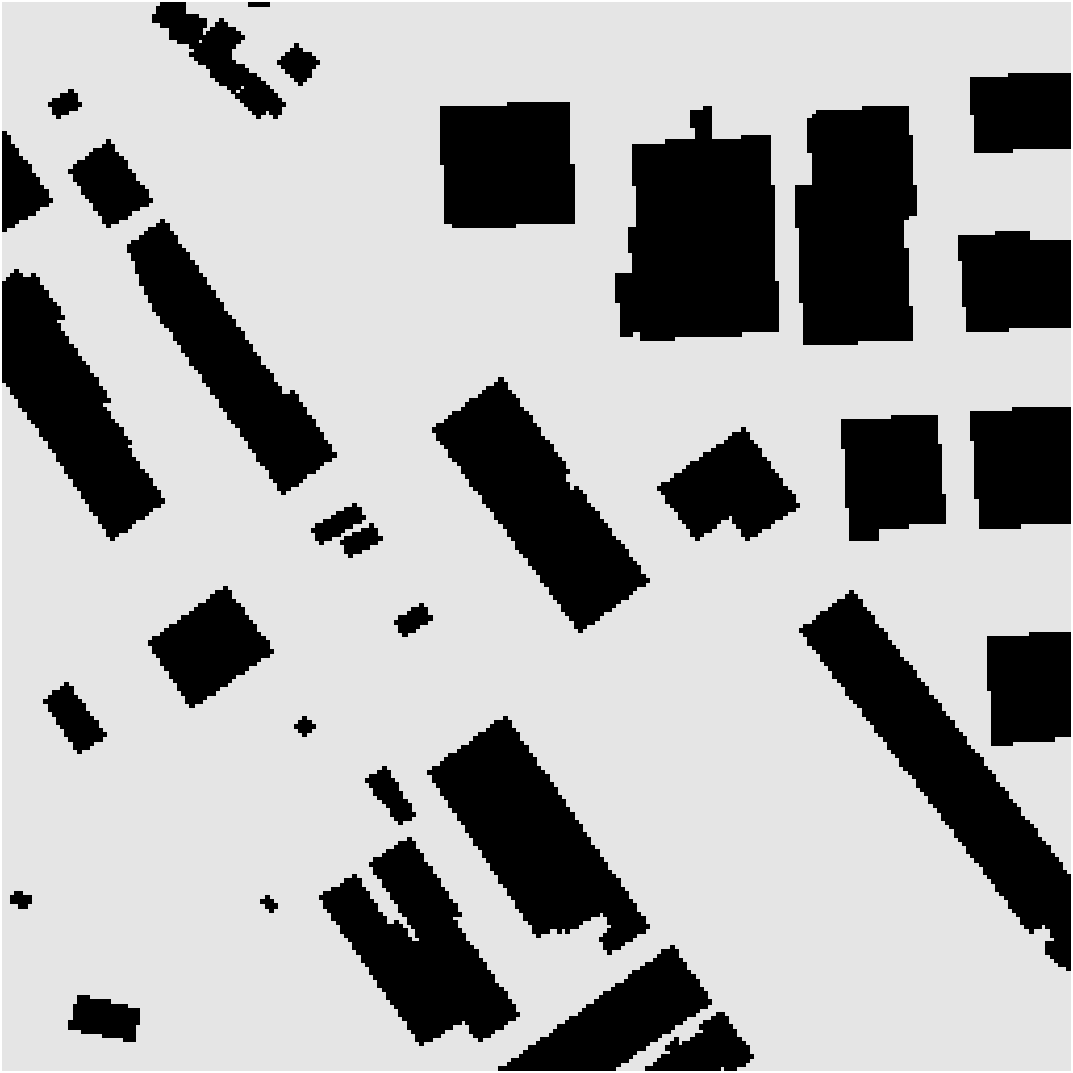}}
  \centerline{C: Boston\_2\_256}
  \centerline{256*256}
\end{minipage}
\hfill
\begin{minipage}{.24\linewidth}
  \centerline{\includegraphics[width=2.1cm]{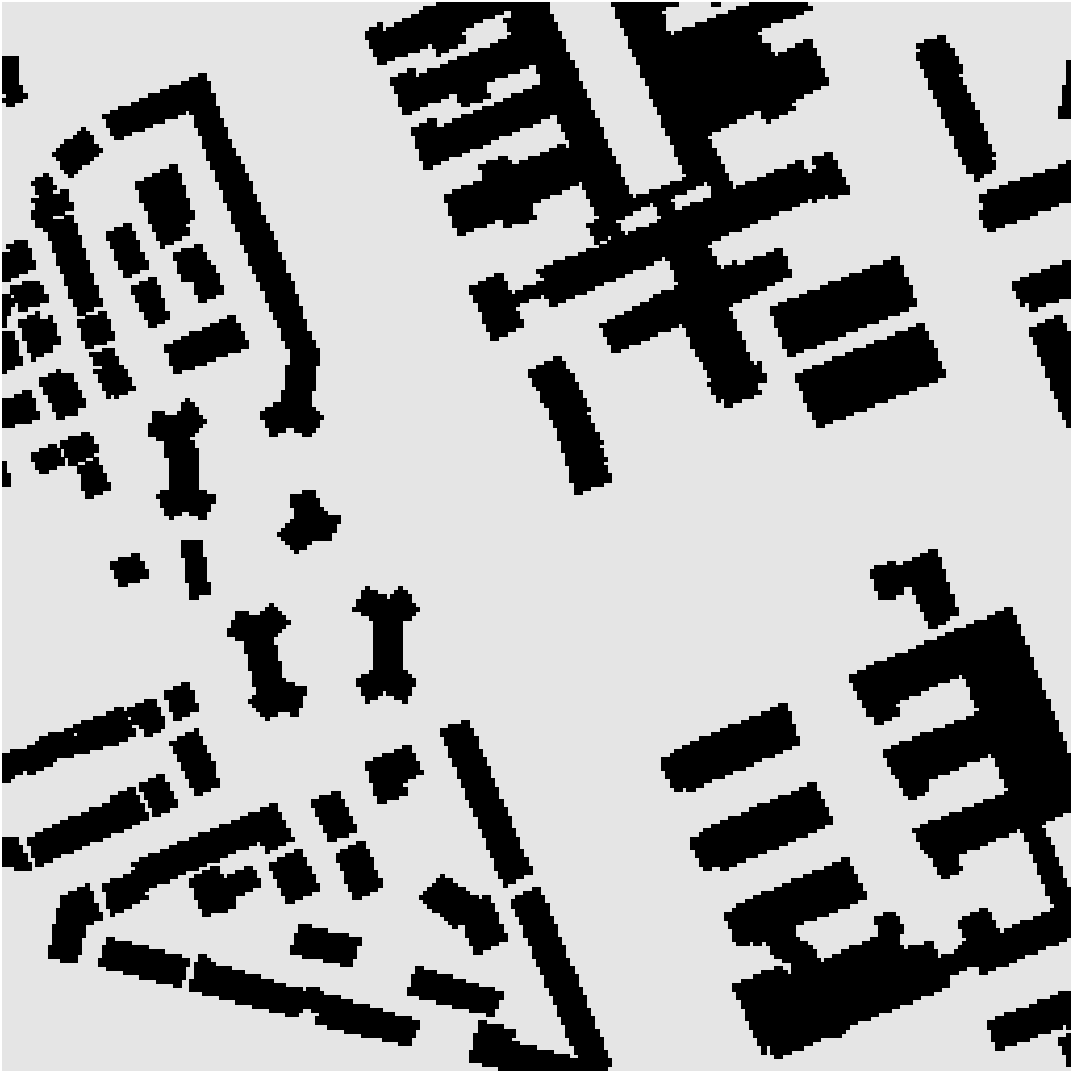}}
  \centerline{D: Milan\_2\_256}
  \centerline{256*256}
\end{minipage}
\vfill
\vspace{0.2cm}
\begin{minipage}{.24\linewidth}
  \centerline{\includegraphics[width=2.1cm]{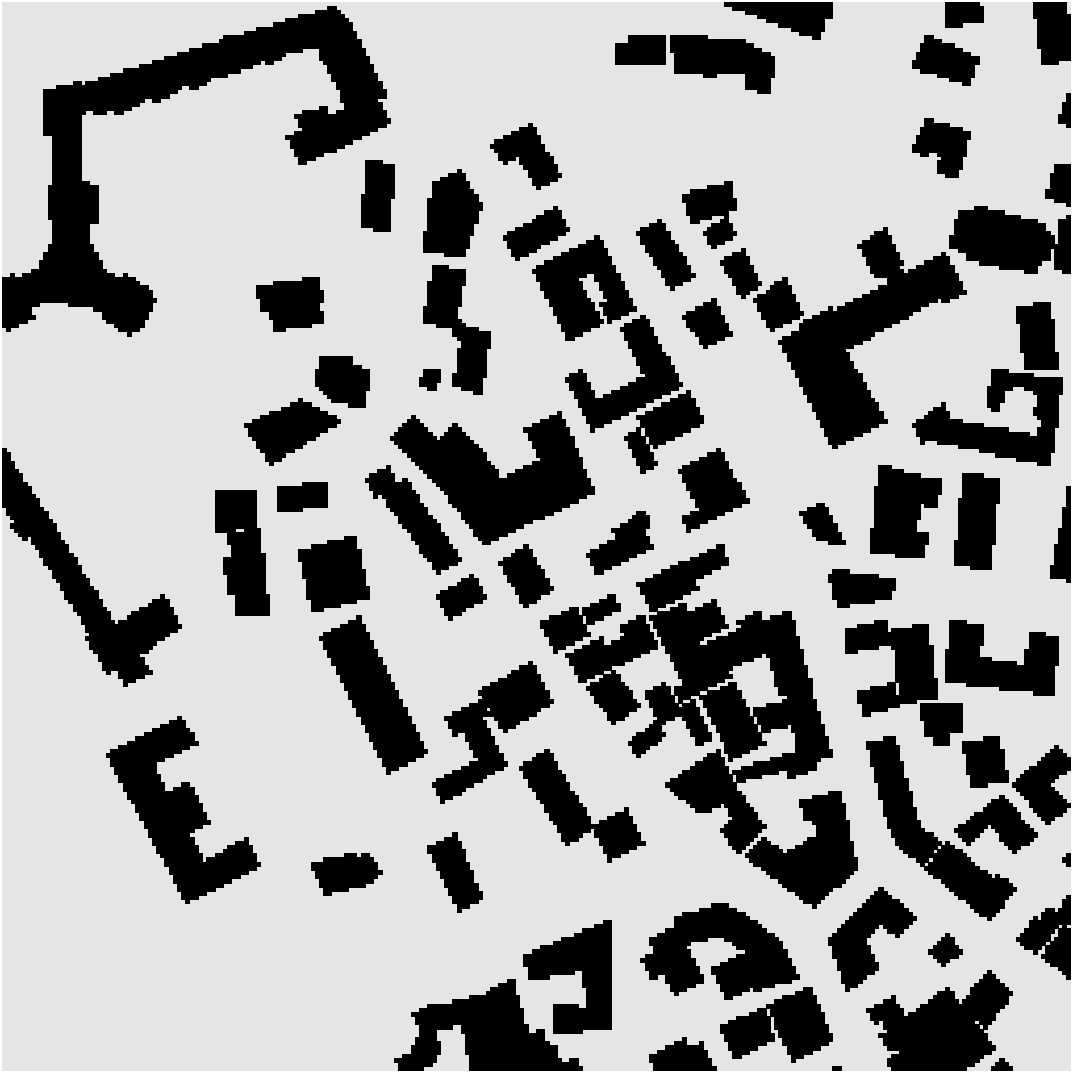}}
  \centerline{E: Moscow\_2\_256}
  \centerline{256*256}
\end{minipage}
\hfill
\begin{minipage}{.24\linewidth}
  \centerline{\includegraphics[width=2.1cm]{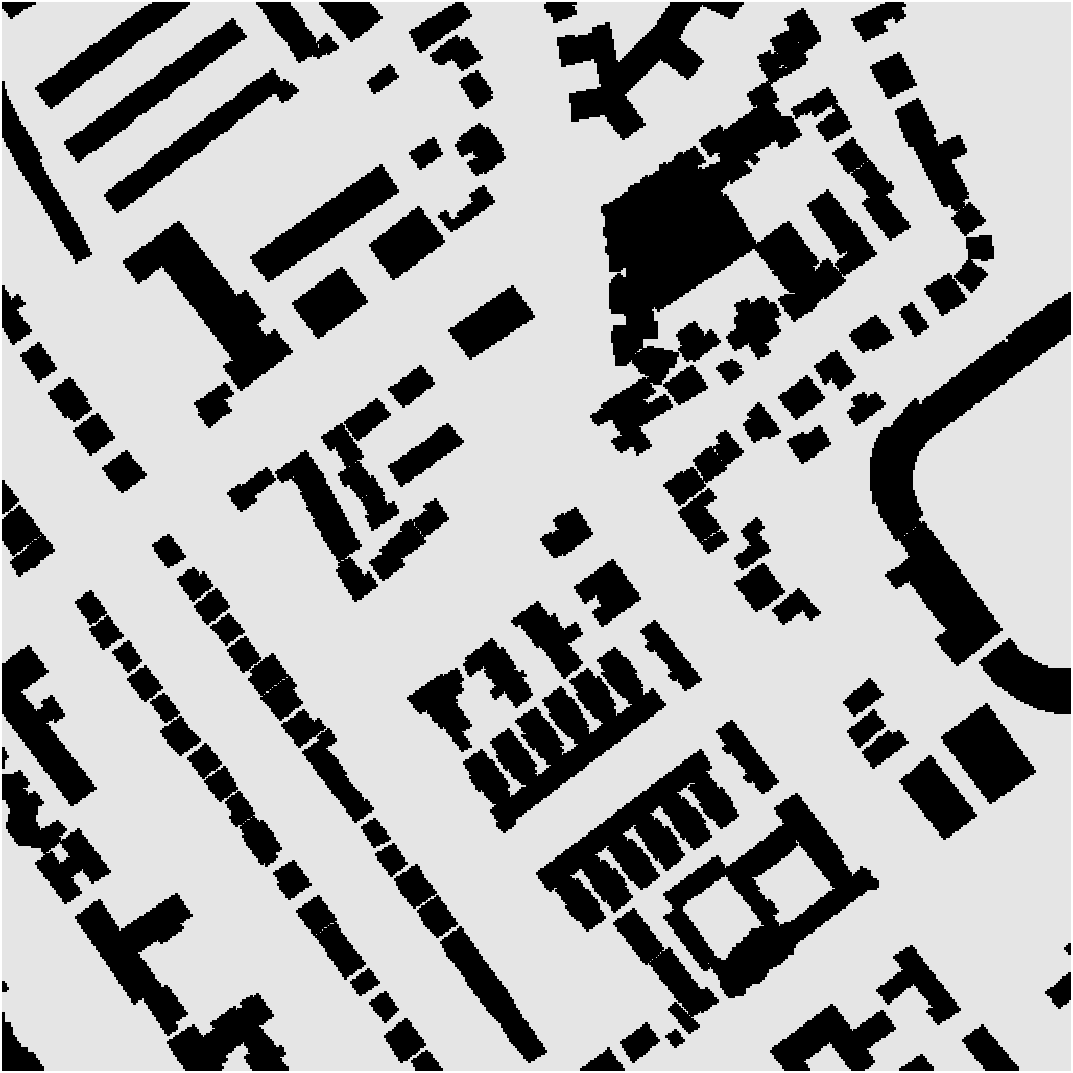}}
  \centerline{F: London\_0\_512}
  \centerline{512*512}
\end{minipage}
\hfill
\begin{minipage}{.24\linewidth}
  \centerline{\includegraphics[width=2.1cm]{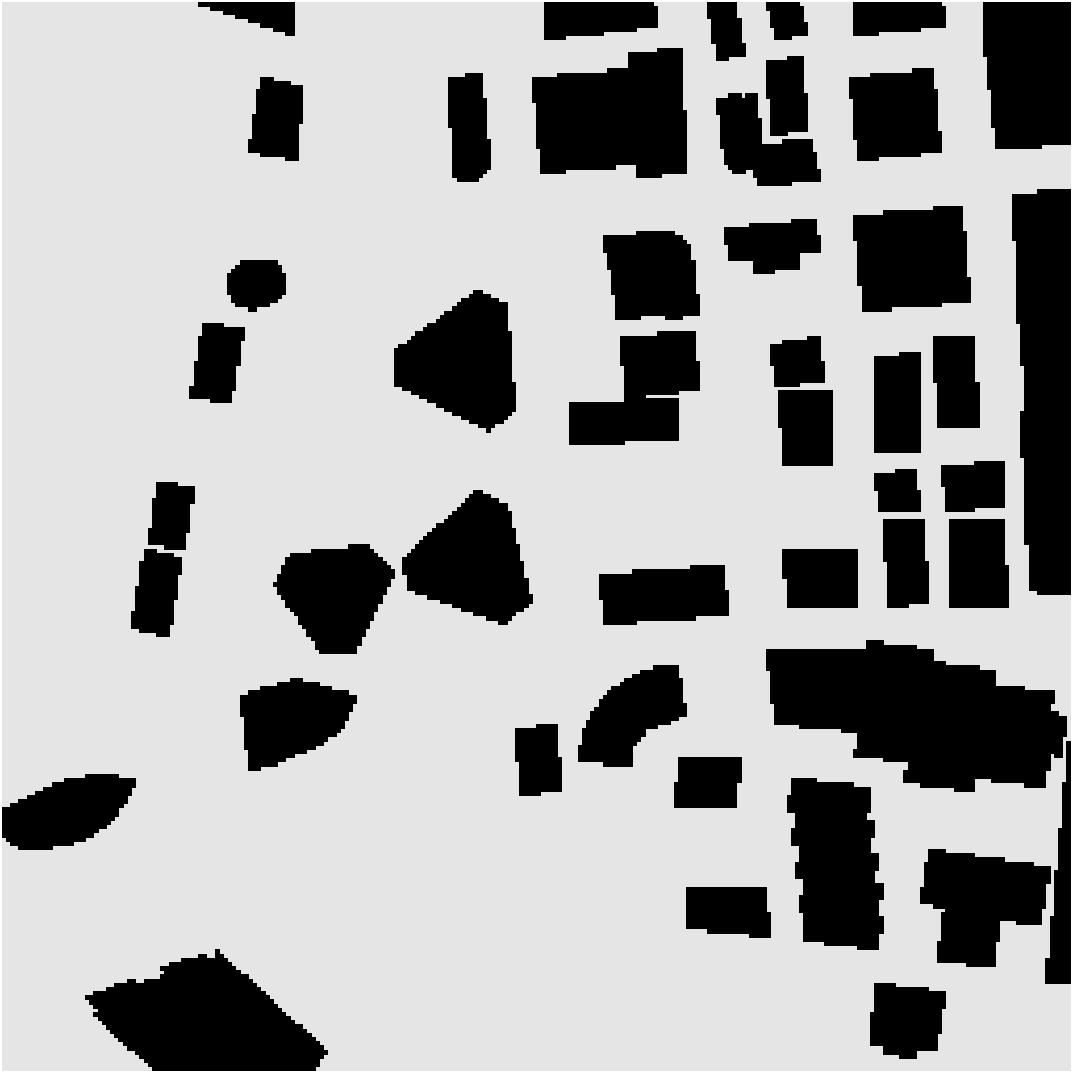}}
  \centerline{G: Sydney\_1\_256}
  \centerline{256*256}
\end{minipage}
\hfill
\begin{minipage}{.24\linewidth}
  \centerline{\includegraphics[width=2.1cm]{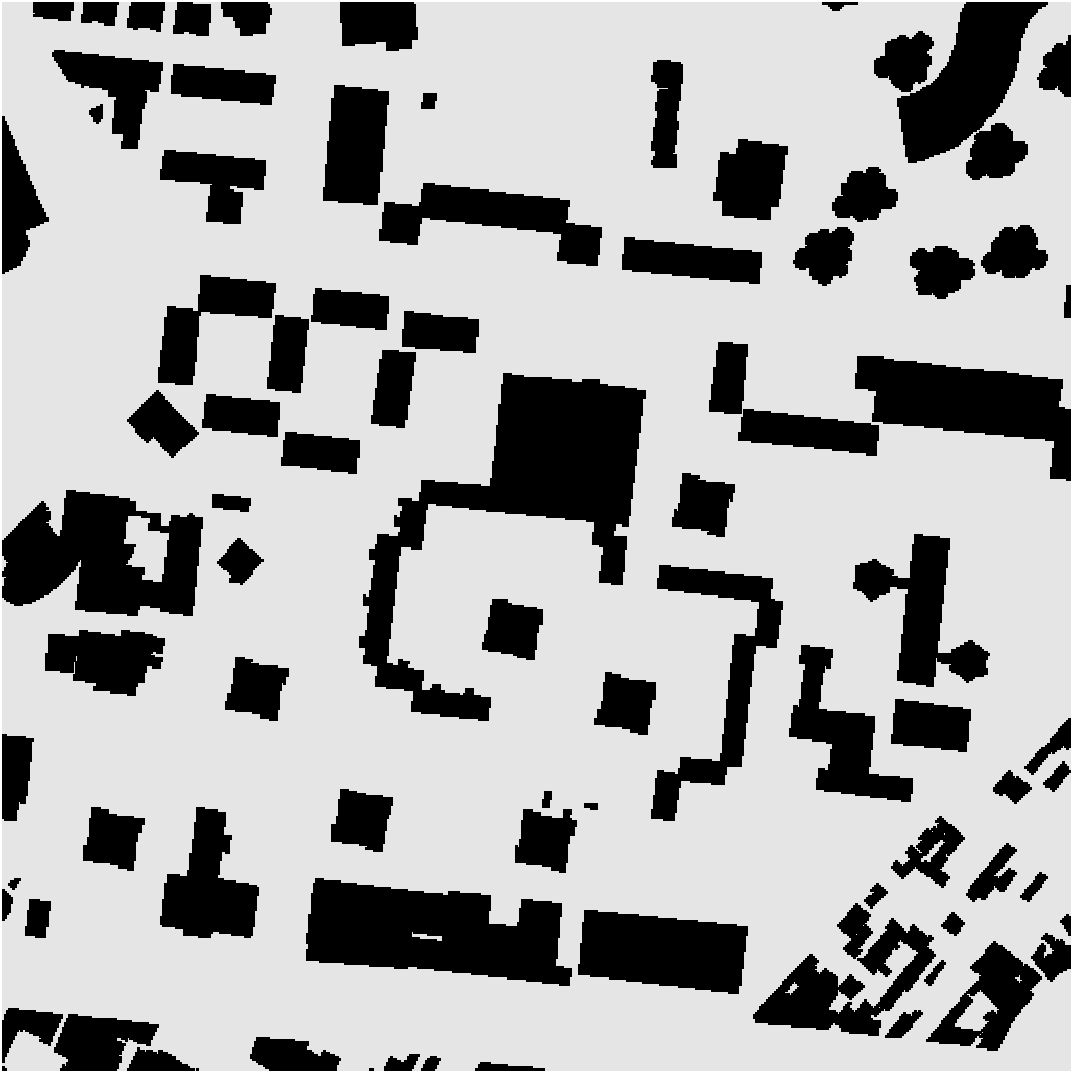}}
  \centerline{H: Paris\_0\_512}
  \centerline{512*512}
\end{minipage}
\vfill

\caption{These figures display eight grid maps used in the comparison with other algorithms. The scale of each map is listed below its name.
}
\label{city_maps}
\end{figure}

If there are not enough unfinished paths in the priority queue, we pop unfinished paths from the secondary queue and insert them into the priority queue until there are $K$ elements in the priority queue, as shown in Algorithm \ref{a11}. We have highlighted the code related to the priority queue.

Essentially, this approach prioritizes the expansion order of unfinished paths, with paths that are closer to the target being expanded earlier than others. This ensures that there is no loss of completeness. An example illustrating the difference between introducing the priority queue and not introducing it is shown in Fig. \ref{priority}. The map used for this comparison is Berlin\_1\_256.map from the public grid map dataset mentioned earlier, with the start and target points set at (59, 72) and (109, 214), respectively.

\begin{table}[t]
	\centering  
	\caption{Details of constructe tangent graph of grid maps}  
	\label{2d_precomputation}  
	\begin{tabular}{|c|c|c|c|c|}  
		\hline  
		Index & MapName & Time Cost(ms) & Nodes & File Size(KB) \\  
		\hline
		1 & Berlin$\_$1$\_$256 & 505.555 &  1631 &  67.1  \\
		\hline
		2 & Denver\_2\_512 & 516.89 &  448  &  17.0 \\
		\hline
		3 & Boston\_2\_256 & 505.588 &  1190  & 103.5 \\
		\hline
		4 & Milan\_2\_256 & 506.5 &  2054   & 106.7  \\
		\hline		
		5 & Moscow\_2\_256 & 506.99 &  2332  & 115.0\\
		\hline
		6 & London\_2\_512 & 3040.1 & 7919  &  828.1  \\
		\hline
		7 & Sydney\_1\_256 & 504.66 &  673 & 28.4 \\
		\hline
		8 & Paris\_0\_512 & 517.76 & 1885 & 142.7 \\
		\hline		
	\end{tabular}
\end{table}


\section{Results}
\label{Results}



In this section, we provide more detailed insights into the performance of our algorithm and compare it with other distinctive topology path planning methods. In the first subsection, we elaborate on the construction of the tangent graph. To ensure a fair comparison, we use the implementations provided by the respective authors. We compare our method with HA*, HTheta*, and RHCF in terms of the total time cost, average time cost for each path, and path length as the required number of paths increases. HA* and HTheta* are H-signature-based algorithms\footnote{https://github.com/subh83/DOSL}, while RHCF is a Voronoi graph-based algorithm\footnote{https://github.com/srl-freiburg/srl\_rhcf\_planner}.

Our experiments are conducted using a well-known grid map dataset\cite{sturtevant2012benchmarks}\footnote{https://movingai.com/benchmarks/grids.html}. To obtain more distinctive topology paths, we select city maps from the dataset, as illustrated in Fig. \ref{city_maps}. For each map, we randomly sample 100 start and target combinations as inputs for path planning. The experiments were carried out on a laptop running Ubuntu 20.04, equipped with a Ryzen 7 5800h (3.2GHz) CPU and 16GB of memory.


\subsection{Construction of tangent graph}

In this section, we provide details on the construction of the tangent graph for the eight maps mentioned. This includes information on the time cost, the number of nodes, and the size of the file used to save the graph, as shown in Table \ref{2d_precomputation}.


The construction of the tangent graph for almost all maps is completed in approximately 0.5 seconds, which is nearly real-time updating. However, London\_2\_512 takes 3 seconds due to its significantly higher number of tangent nodes compared to other maps, roughly three times more. The file size required to save the tangent graph for these maps ranges from 60KB to 830KB, which is manageable for an ordinary platform.

\subsection{Comparison with other methods}

In this section, we focus on comparing our method's efficiency with other methods in terms of time cost. Specifically, we analyze how the time cost changes as the number of required paths increases. This analysis includes the total cost and the average time cost for one path under the eight maps mentioned in the previous section. It's worth noting that we have set an upper time bound of 10 seconds for all methods, as some algorithms may exceed this threshold when attempting to find hundreds of topology distinctive paths.

\begin{figure}[t]
\centering
\vspace*{8pt}
\includegraphics[width=8.cm]{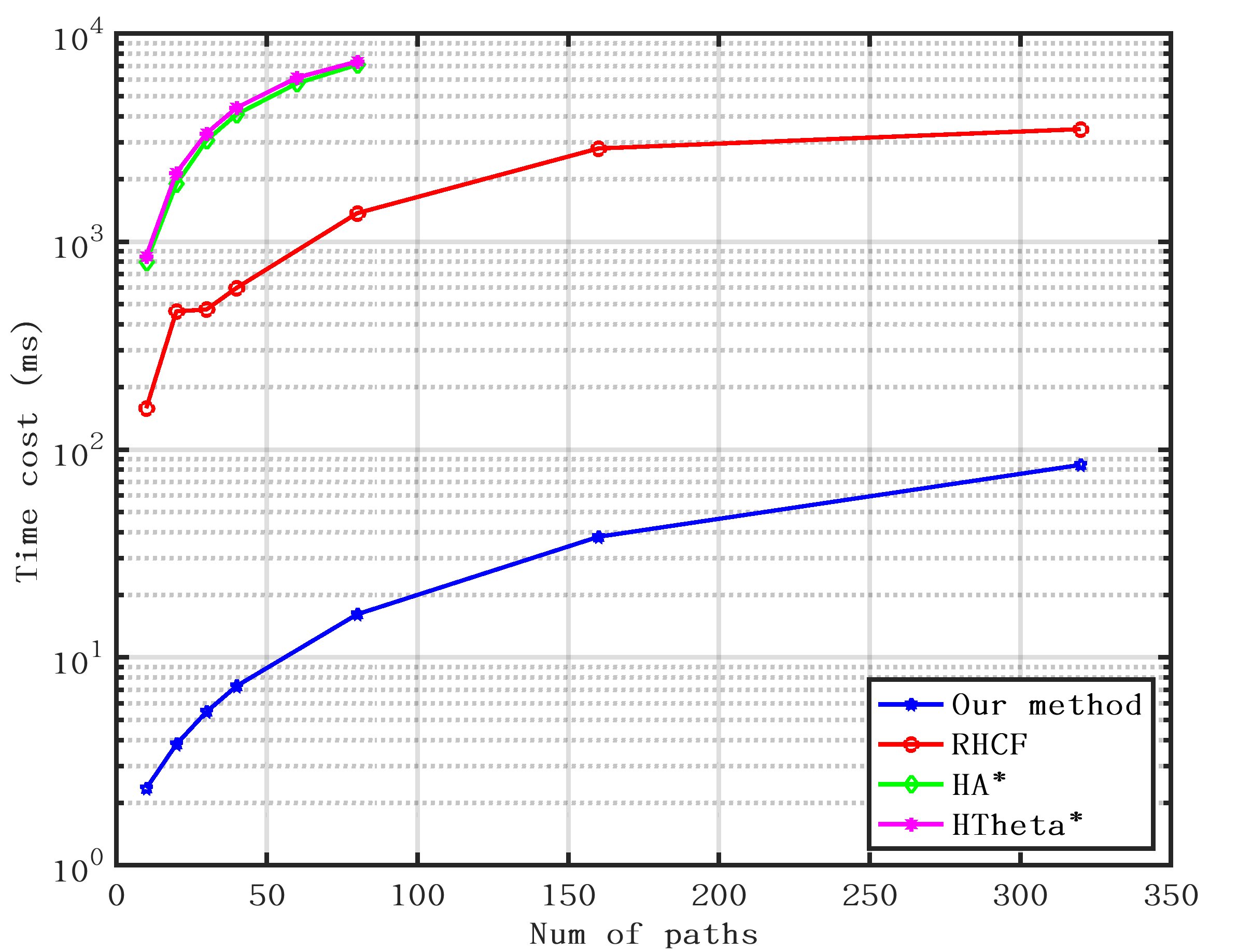}
\caption{The figure the mean time cost of the four methods search multiple paths under mentioned the eight maps. Due to the difference in the order of magnitude of the total time cost of various methods, the vertical axis of the chart is logarithmic. } 
\label{mean_time_cost_comparison}
\end{figure}

\begin{figure}[t]
\centering
\vspace*{8pt}
\includegraphics[width=8.cm]{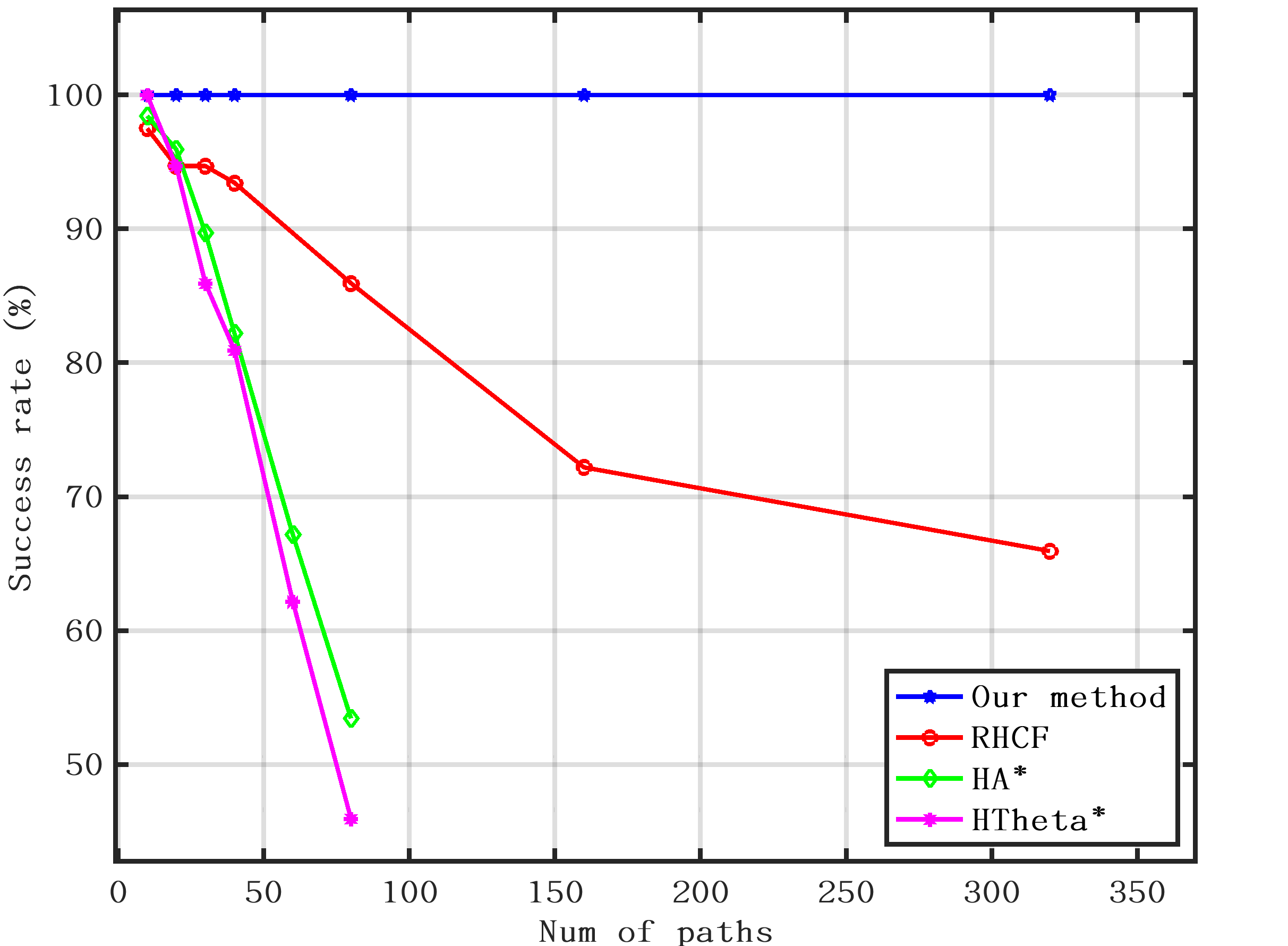}
\caption{
The figure shows the mean time cost of the four methods when searching for multiple paths under the eight maps mentioned. Due to the difference in the order of magnitude of the total time cost for various methods, the vertical axis of the chart is logarithmic.} 
\label{mean_suc_rate_comparison}
\end{figure}

Specifically, we configured our method and RHCF to find 10, 20, 30, 40, 80, 160, and 320 paths, as they require relatively low time costs. Additionally, we set HA* and HTheta* to find 10, 20, 30, 40, 60, and 80 paths, as searching for more paths would exceed the 10-second time limit. The mean time costs under various maps are presented in Fig. \ref{mean_time_cost_comparison}, and the success rate (finding the required paths within 10 seconds) of each method is shown in Fig. \ref{mean_suc_rate_comparison}.


\begin{figure}[t]
\centering
\vspace*{8pt}
\includegraphics[width=8.cm]{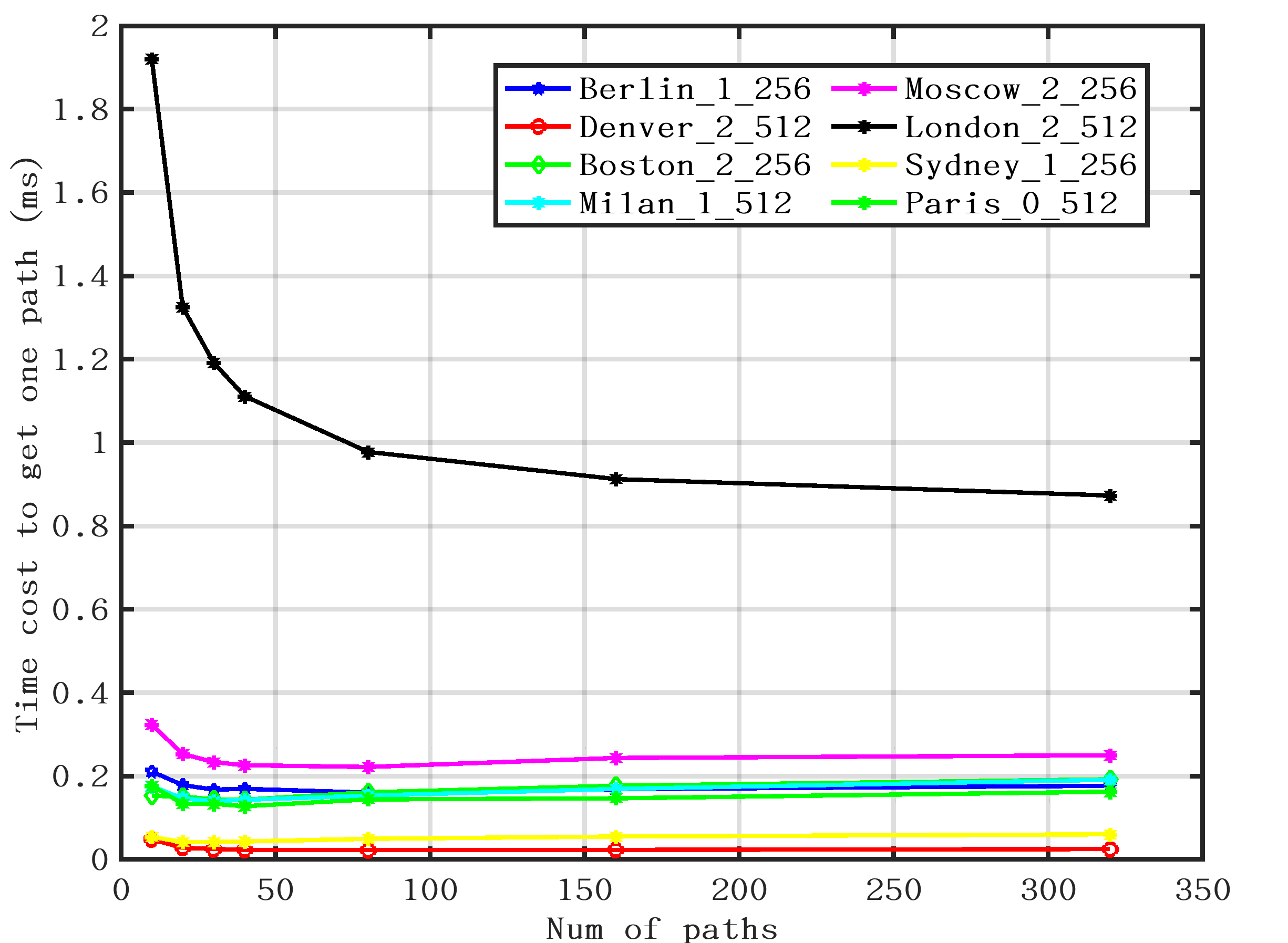}
\caption{
The figure demonstrates how the mean time cost of our method to search for a single path changes as the total number of paths and the map used vary.} 
\label{mean_path_time_cost}
\end{figure}

As shown in Fig. \ref{mean_time_cost_comparison}, our method exhibits the lowest time cost, averaging 84ms to find 320 paths, which is smaller in magnitude compared to other methods. Consequently, in terms of success rate, our method achieves a 100\% success rate in finding the required paths, while the success rates of other methods noticeably decrease as the number of paths increases.

Next, we investigated how the time cost to determine a single path changes for our method as the number of paths increases under the eight maps, as depicted in Fig. \ref{mean_path_time_cost}. The mean cost of our method to obtain one path initially decreases and then stabilizes as the total number of paths increases. This trend is more pronounced in more complex maps. This effect occurs because our method limits the number of nodes to be expanded during each iteration to no more than the total number of paths.




\section{Discussion and conclusion}
\label{Conclusion}


In this article, we introduce a tangent graph-based topologically distinctive path planning algorithm. It leverages the property that tangents form locally shortest paths, ensuring that any two locally shortest paths belong to different topologies. Compared to existing algorithms, our approach requires no indicator to determine whether two paths belong to the same topology. Furthermore, it eliminates the need to repeat the search for multiple paths, as all distinctive paths can be found in a single search.

To address the challenge of the exponential growth in queue size during breadth-first search, we propose a priority limitation technique. This approach significantly reduces the time cost of searching for multiple paths while preserving computational complexity.


Our method consists of two main steps: the construction of the tangent graph and the search for multiple paths. In the construction of the tangent graph step, we employ the locally collide constraint on edges to ensure that each segment is locally the shortest. Additionally, to avoid repeating these calculations every time the map is loaded, we save the results to a binary file, which is then loaded during the framework loading process, saving valuable time.

During the search for multiple paths, we introduce a ``no-loop" constraint to prevent the repetition of waypoints in the path and avoid intersections with itself. Additionally, we apply the ``get closer to obstacle" ETC to ensure locally the shortest path. It's important to note that as the number of waypoints in the path increases, the time cost of loop detection also increases. Experimental results show that the ratio of no-loop constraint checks in the total cost increases as the required number of paths increases. This occurs because the requirement for more paths often results in paths with a larger number of waypoints.


In comparison with other methods, our approach demonstrates significant efficiency when compared to RHCF, HA*, and HTheta*. Specifically, our method takes approximately 100ms to determine 320 paths from the mentioned public map dataset, while RHCF takes more than 1 second, and HA* and HTheta* take more than 10 seconds. However, it's important to note that our method has a limitation in that it cannot search for multiple paths with the same topology, unlike methods that use indicators like H-signature, which are capable of searching for multiple paths with the same topology.





In the future, we plan to implement our method using multiple threads, as currently, it operates in a single thread. Since there is no dependence between different unfinish paths during the search process, there will be no loss of complexity when utilizing multiple threads. Additionally, we intend to apply our method to trajectory optimizations. Given that our method can provide hundreds of topology-distinctive paths in real time, it is crucial to study how to speed up trajectory optimizations to enable them to efficiently handle hundreds of paths as input.

\section*{Acknowledgments}
The first author thanks Lu Zhu for her encouragement and
support during the consummation of this work.

\normalem
\printbibliography

@article{rosmann2017integrated,
  title={Integrated online trajectory planning and optimization in distinctive topologies},
  author={R{\"o}smann, Christoph and Hoffmann, Frank and Bertram, Torsten},
  journal={Robotics and Autonomous Systems},
  volume={88},
  pages={142--153},
  year={2017},
  publisher={Elsevier}
}

@inproceedings{ranganeni2018effective,
  title={Effective footstep planning for humanoids using homotopy-class guidance},
  author={Ranganeni, Vinitha and Salzman, Oren and Likhachev, Maxim},
  booktitle={Proceedings of the International Conference on Automated Planning and Scheduling},
  volume={28},
  pages={500--508},
  year={2018}
}

@inproceedings{kim2021topology,
  title={Topology-guided path planning for reliable visual navigation of MAVs},
  author={Kim, Dabin and Kim, Gyeong Chan and Jang, Youngseok and Kim, H Jin},
  booktitle={2021 IEEE/RSJ International Conference on Intelligent Robots and Systems (IROS)},
  pages={3117--3124},
  year={2021},
  organization={IEEE}
}

@inproceedings{bhattacharya2010search,
  title={Search-based path planning with homotopy class constraints},
  author={Bhattacharya, Subhrajit},
  booktitle={Proceedings of the AAAI conference on artificial intelligence},
  volume={24},
  number={1},
  pages={1230--1237},
  year={2010}
}

@inproceedings{bhattacharya2012search,
  title={Search-based path planning with homotopy class constraints in 3D},
  author={Bhattacharya, Subhrajit and Likhachev, Maxim and Kumar, Vijay},
  booktitle={Proceedings of the AAAI Conference on Artificial Intelligence},
  volume={26},
  number={1},
  pages={2097--2099},
  year={2012}
}

@article{bhattacharya2012topological,
  title={Topological constraints in search-based robot path planning},
  author={Bhattacharya, Subhrajit and Likhachev, Maxim and Kumar, Vijay},
  journal={Autonomous Robots},
  volume={33},
  pages={273--290},
  year={2012},
  publisher={Springer}
}

@inproceedings{kim2012optimal,
  title={Optimal trajectory generation under homology class constraints},
  author={Kim, Soonkyum and Sreenath, Koushil and Bhattacharya, Subhrajit and Kumar, Vijay},
  booktitle={2012 IEEE 51st IEEE Conference on Decision and Control (CDC)},
  pages={3157--3164},
  year={2012},
  organization={IEEE}
}

@inproceedings{kuderer2014online,
  title={Online generation of homotopically distinct navigation paths},
  author={Kuderer, Markus and Sprunk, Christoph and Kretzschmar, Henrik and Burgard, Wolfram},
  booktitle={2014 IEEE International Conference on Robotics and Automation (ICRA)},
  pages={6462--6467},
  year={2014},
  organization={IEEE}
}

@inproceedings{yi2017topology,
  title={Topology-aware RRT* for parallel optimal sampling in topologies},
  author={Yi, Daqing and Goodrich, Michael A and Howard, Thomas M and Seppi, Kevin D},
  booktitle={2017 IEEE International Conference on Systems, Man, and Cybernetics (SMC)},
  pages={513--518},
  year={2017},
  organization={IEEE}
}

@article{palmieri2015fast,
  title={A fast randomized method to find homotopy classes for socially-aware navigation},
  author={Palmieri, Luigi and Rudenko, Andrey and Arras, Kai O},
  journal={arXiv preprint arXiv:1510.08233},
  year={2015}
}

@article{yen1971finding,
  title={Finding the k shortest loopless paths in a network},
  author={Yen, Jin Y},
  journal={management Science},
  volume={17},
  number={11},
  pages={712--716},
  year={1971},
  publisher={Informs}
}

@article{rice2020multi,
  title={Multi-homotopy class optimal path planning for manipulation with one degree of redundancy},
  author={Rice, Jacob J and Schimmels, Joseph M},
  journal={Mechanism and Machine Theory},
  volume={149},
  pages={103834},
  year={2020},
  publisher={Elsevier}
}

@article{yao2021tangent,
  title={Tangent Based-Path Path Planning with Distinctive Topologies in a Dynamic Environment},
  author={Yao, Zhuo},
  year={2021},
  publisher={Preprints},
  organization={Preprints}
}

@article{liu1991proposal,
  title={Proposal of tangent graph and extended tangent graph for path planning of mobile robots},
  author={Liu, Y-H and Arimoto, Suguru},
  booktitle={Proceedings. 1991 IEEE International Conference on Robotics and Automation},
  pages={312--317},
  year={1991},
  organization={IEEE}
}

@article{liu1992path,
  title={Path planning using a tangent graph for mobile robots among polygonal and curved obstacles: Communication},
  author={Liu, Yun-Hui and Arimoto, Suguru},
  journal={The International Journal of Robotics Research},
  volume={11},
  number={4},
  pages={376--382},
  year={1992},
  publisher={Sage Publications Sage CA: Thousand Oaks, CA}
}

@article{sturtevant2012benchmarks,
  title={Benchmarks for Grid-Based Pathfinding},
  author={Sturtevant, N.},
  journal={Transactions on Computational Intelligence and AI in Games},
  volume={4},
  number={2},
  pages={144 -- 148},
  year={2012},
  url = {http://web.cs.du.edu/~sturtevant/papers/benchmarks.pdf},
}

@article{yao2019reinforcedrimjump,
  title={ReinforcedRimJump: Tangent-based shortest-path planning for two-dimensional maps},
  author={Yao, Zhuo and Zhang, Weimin and Shi, Yongliang and Li, Mingzhu and Liang, Zhenshuo and Huang, Qiang},
  journal={IEEE Transactions on Industrial Informatics},
  year={2019},
  publisher={IEEE}
}

@inproceedings{oh2017edge,
  title={Edge n-level sparse visibility graphs: Fast optimal any-angle pathfinding using hierarchical taut paths},
  author={Oh, Shunhao and Leong, Hon Wai},
  booktitle={Tenth Annual Symposium on Combinatorial Search},
  year={2017}
}

\end{document}